\documentclass[preprint,12pt]{elsarticle}

\usepackage{amssymb}
\usepackage{amsmath}

\usepackage{amssymb}
\usepackage{diagbox} 
\usepackage{graphicx}
\usepackage{amsmath} 
\usepackage{amssymb}
\usepackage{booktabs}
\usepackage{color}
\usepackage{subfigure}
\usepackage{mathrsfs}
\usepackage{multirow}
\usepackage{amsthm}
\usepackage{algorithm}
\usepackage{algorithmic}

\usepackage{url}
\usepackage{hyperref}

\usepackage{makecell}
\usepackage{array} 
\usepackage{bbding}
\usepackage{pifont}
\usepackage{wasysym}
\usepackage{utfsym}
\usepackage[section]{placeins}
\usepackage{textcomp}
\usepackage{enumitem}

\newtheorem{theorem}{Theorem}
\newtheorem{proposition}[theorem]{Proposition}
\newtheorem{lemma}{Lemma}

\newtheorem{definition}{Definition}
\newtheorem{assumption}{Assumption}
\newtheorem{remark}{Remark}

\journal{Arxiv}

\begin{document}

\begin{frontmatter}



\title{Meta Additive Model: Interpretable Sparse Learning With Auto Weighting}

\author[label1]{Xuelin Zhang} 
\ead{zhangxuelin@webmail.hazu.edu.cn}
\author[label1]{Xinyue Liu} 
\ead{lxy36@webmail.hzau.edu.cn}

\author[label1]{Lingjuan Wu} 
\ead{wulj@mail.hzau.edu.cn}
\author[label1]{Hong Chen\corref{cor1}} 
\ead{chenh@mail.hzau.edu.cn}

\cortext[cor1]{Corresponding author.}
\affiliation[label1]{organization={College of Informatics, Huazhong Agricultural University},
            city={Wuhan},
            country={China}}

\begin{abstract}
Sparse additive models have attracted much attention in high-dimensional data analysis due to their flexible representation and strong interpretability. However, most existing models are limited to single-level learning under the mean-squared error criterion, whose empirical performance can degrade significantly in the presence of complex noise, such as non-Gaussian perturbations, outliers, noisy labels, and imbalanced categories. The sample reweighting strategy is widely used to reduce the model's sensitivity to atypical data; however, it typically requires prespecifying the weighting functions and manually selecting additional hyperparameters.  
To address this issue, we propose a new \emph{meta additive model} (MAM) based on the bilevel optimization framework, which learns data-driven weighting of individual losses by parameterizing the weighting function via an MLP trained on meta data. 
MAM is capable of a variety of learning tasks, including variable selection, robust regression estimation, and imbalanced classification. 
Theoretically, MAM provides guarantees on convergence in computation, algorithmic generalization, and variable selection consistency under mild conditions. Empirically, MAM outperforms several state-of-the-art additive models on both synthetic and real-world data under various data corruptions. 
\end{abstract}



\end{frontmatter}


\section{Introduction} \label{section1}

Let $\mathcal{X} \subset \mathbb{R}^p$ be an input space and let $\mathcal{Y} \subset \mathbb{R}$ be the corresponding output set. Assume that random observation  $(X,Y)\in\mathcal{X}\times\mathcal{Y}$ is  generated by 
\begin{equation} \label{data-model}
Y=f^*(X)+\epsilon,
\end{equation}
where $f^*$ stands for the unknown target function and $\epsilon$ represents the random noise. 
Usually, $f^*$ can be  approximated by data-driven models associated with empirical observations $\{(x_i, y_i)\}_{i=1}^n \subset \mathcal{X}\times \mathcal{Y}$ generated from Eq.\eqref{data-model}.
However, in the high-dimensional input setting, nonparametric models often exhibit low convergence rates due to the so-called \emph{curse of dimensionality} \citep{stone1985additive}. To address this issue, various additive models have been formulated for different data circumstances and learning tasks \citep{wang2023tilted,duong2024cat}. It is well known that additive models possess attractive properties, including the ability to overcome the curse of dimensionality, the flexibility of function approximation, and the capacity to select variables \citep{kandasamy2016additive,agarwal2021neural}.

Additive models require that the hypothesis space has an additive structure to enable interpretable predictions. One typical way is to
directly decompose the input space $\mathcal{X}$ into $p$ disjoint parts $\{\mathcal{X}_j\}_{j=1}^p$ with $\mathcal{X}_j \subset \mathbb{R}$, $j\in\{1,...,p\}$. Naturally,  additive hypothesis space $\mathcal{H}$ can be represented as
\begin{equation}
\mathcal{H} =\big\{f: f(X)=\sum_{j=1}^p f_j(X_j), f_j\in\mathcal{H}_j, X_j\in\mathcal{X}_j\big\},
\end{equation}
where $\mathcal{H}_j$ is the component function space on $\mathcal{X}_j$. 
The candidates of component function spaces include the basis expansion space \citep{chen2020sparse,wang2021huber}, the reproducing kernel Hilbert space (RKHS) \citep{christmann2016learning,lu2022additive,wang2023tilted}, and the neural networks-based space \citep{agarwal2021neural,duong2024cat}.

Along with statistical modeling, theoretical properties and empirical evaluations of additive models have been extensively investigated under the classical empirical risk minimization framework, i.e., $\min_{f \in \mathcal{H}} \sum_{i=1}^n \ell\left(y_i,f\left(x_i\right)\right)$, where $\ell(\cdot)$ is a loss function. However, most existing approaches are formulated based on the squared loss $\ell(f(x), y)=(f(x)-y)^2$ for regression tasks \citep{raskutti2012minimax,16} and the hinge loss (or logistic loss) for classification tasks \citep{liu2007spam}, which often perform poorly in the presence of training data bias, e.g., training data corrupted by non-Gaussian noise, noisy and imbalanced labels. To improve the algorithmic robustness, several approaches of additive models have been proposed by pre-specifying loss functions, e.g., the quantile loss \citep{lv2018oracle}, the mode-induced risk \citep{chen2020sparse}, and Huber loss \citep{wang2021huber} for regression as well as the ramp loss \citep{chen2021sparse}
and the correntropy-induced loss \citep{yuan2023sparse} for classification. 
Usually, for robust estimation, atypical samples (e.g., outliers or data with skewed noise) should be suppressed by assigning lower weights during training. On the contrary, when dealing with imbalanced data, samples with larger loss values should be emphasized, as they are likely complex samples near the classification decision boundary. 
Clearly, the prespecified loss functions may not exhibit the adaptivity and flexibility to handle diverse data conditions. In particular, it is challenging to manually determine the additional hyperparameters of the robust loss \citep{shu2019meta, shu2023learning}. 

Recently, a robust sparse additive model was proposed in \citep{wang2023tilted} under the principle of tilted empirical risk minimization (TERM) \citep{li2020tilted}, which simultaneously enables variable selection, robust estimation, and imbalanced classification by flexibly assigning weights to each sample. However, this framework is heavily dependent on the choice of TERM-induced hyperparameter, where the negative hyperparameter helps improve noise-tolerant robustness, and the positive one achieves imbalanced classification \citep{li2020tilted}. To achieve multi-objective classification, a nested TERM framework is developed by grouping training samples with their labels and introducing an additional hyperparameter, which significantly increases the computational cost of hyperparameter selection.

To address the aforementioned issues, we integrate meta-learning into sparse additive models and formulate a new meta-additive model (MAM) to enable auto-weighting and sparse approximation. In contrast to the previous robust additive models, our key idea is to additionally use an MLP network as the weighting function, whose parameters are automatically updated by the bilevel optimization framework. In addition to empirical evaluations based on extensive experimental results, we provide theoretical foundations for the computational convergence of our training algorithm, the upper bounds for generalization errors, and the statistical consistency of variable selection. 

The main contributions of this paper can be summarized as follows:
\begin{itemize}
\item  \emph{Bilevel statistical modeling.} To the best of our knowledge, our MAM is the first meta learning method for additive models, where the formulated bilevel optimization scheme can tackle data-driven weighting, robust estimation, and sparse variable selection simultaneously. Additionally, the proposed MAM can handle imbalanced classification with contaminated data and multi-objective learning.  

\item \emph{Theoretical guarantees}. Under conditions similar to those in \citep{shu2023cmw,shu2023learning}, we present convergence and generalization guarantees for MAM. In particular, the proposed MAM achieves variable selection consistency, extending the previous single-level results \citep{chen2020sparse,wang2023tilted} to complex bilevel optimization settings.

\item \emph{Empirical competitiveness}. The experimental evaluations on both synthetic and real-world data demonstrate that MAM can successfully identify truly informative variables, and provide interpretable and robust predictions, particularly in challenging scenarios involving corrupted or imbalanced data.
\end{itemize}

\begin{table*}[!t]
\centering 
\caption{Summary of properties for MAM and its related models ($\checkmark$ = enjoying the information, $\times$ = not available, "ECN" stands for "empirical covering number", $T$ denotes the maximum  number of iterations, $m$ stands for the data size, and the constant $c\in (0,1)$.)}
\resizebox{\textwidth}{!}{
\begin{tabular}{c|cccccccccc}
\hline
\textbf{Property $\backslash$ Model}&\centering \textbf{SpAM} \citep{liu2007spam} &\centering \textbf{SAQR} \citep{lv2018oracle} &\centering \textbf{SpMAM} \citep{chen2020sparse} &\centering \textbf{CSAM} \citep{yuan2023sparse}  &\centering \textbf{GIW} \citep{fang2024generalizing} &\centering \textbf{KAN} \citep{liu2025kan} \tabularnewline\hline
\centering Interpretability
&\centering $\checkmark$ &\centering $\checkmark$ &\centering  $\checkmark$&\centering  $\checkmark$ &\centering  $\times$&\centering  $\checkmark$
\tabularnewline
\centering Variable Selection
&\centering $\checkmark$ &\centering $\checkmark$ &\centering  $\checkmark$&\centering  $\checkmark$ &\centering  $\times$&\centering  $\checkmark$
\tabularnewline
\centering  Robust Estimation
&\centering $\times$ &\centering  $\checkmark$ &\centering $\checkmark$ &\centering $\checkmark$ &\centering $\checkmark$ &\centering $\times$
\tabularnewline
\centering  Imbalanced Classification
&\centering $\times$ &\centering  $\times$ &\centering $\times$ &\centering $\times$ &\centering $\checkmark$ &\centering $\times$ 
\tabularnewline
\centering  Multi-objective Learning
&\centering $\times$ &\centering  $\times$ &\centering $\times$ &\centering $\times$ &\centering $\checkmark$ &\centering $\times$
\tabularnewline
\centering Convergence Analysis
&\centering $\times$  & \centering $\checkmark$ & \centering $\times$ & \centering $\times$ & \centering $\times$ & \centering $\times$
\tabularnewline
\centering Variable Selection Analysis
&\centering $\checkmark$  & \centering $\checkmark$ & \centering $\checkmark$ & \centering $\checkmark$ & \centering $\times$ & \centering $\times$
\tabularnewline	
\centering Learning Framework
&\centering Single-level &\centering Single-level &\centering Single-level &\centering Single-level &\centering Single-level &\centering Single-level
\tabularnewline	
\centering Generalization Technique
&\centering $\times$  & \centering $\times$ & \centering $l_2$-ECN & \centering $l_2$-ECN & \centering $\times$ & \centering $\times$
\tabularnewline
\centering Generalization Bounds
&\centering $\times$  & \centering $\times$ & \centering $\mathcal{O}(m^{-2/5})$ & \centering $\mathcal{O}(m^{-1/4})$ & \centering $\times$ & \centering $\times$ \tabularnewline
\centering Regularization
&\centering $\ell_{2,1}$-norm &\centering $\ell_{2,1}$-norm  &\centering $\ell_{q,1}$-norm &\centering $\ell_{q,1}$-norm & \centering $\times$ & \centering $\times$
\tabularnewline	
\centering Loss Function Setting
&\centering Manual & \centering Manual & \centering Manual & \centering Manual \centering Manual & \centering Manual 
\tabularnewline	
\hline
\textbf{Property $\backslash$ Model}&\centering \textbf{TSpAM} \citep{wang2023tilted} &\centering \textbf{CAT} \cite{duong2024cat} &\centering \textbf{NAM} \cite{agarwal2021neural} &\centering \textbf{PBCS} \citep{zhou2022core} &\centering \textbf{MWNet} \citep{shu2019meta}  &\centering \textbf{MAM (Ours)}\tabularnewline\hline
\centering Interpretability
&\centering $\checkmark$ &\centering $\checkmark$ &\centering  $\checkmark$&\centering  $\times$ &\centering  $\times$&\centering  $\checkmark$
\tabularnewline
\centering Variable Selection
&\centering $\checkmark$ &\centering $\times$ &\centering $\checkmark$ &\centering $\checkmark$ &\centering $\times$  &\centering $\checkmark$\tabularnewline
\centering  Robust Estimation
&\centering $\checkmark$ &\centering $\times$ &\centering $\times$ &\centering $\checkmark$ &\centering $\checkmark$ &\centering $\checkmark$\tabularnewline
\centering  Imbalanced Classification
&\centering $\checkmark$ &\centering $\times$ &\centering $\times$ &\centering $\checkmark$ &\centering $\checkmark$  &\centering $\checkmark$\tabularnewline
\centering  Multi-objective Learning
&\centering $\checkmark$ &\centering $\times$ &\centering $\times$ &\centering $\checkmark$ &\centering $\checkmark$  &\centering $\checkmark$\tabularnewline
\centering Convergence Analysis
& \centering $\times$ &\centering $\times$ &\centering $\times$  & \centering $\checkmark$ & \centering $\checkmark$ &\centering $\checkmark$\tabularnewline
\centering Variable Selection Analysis
& \centering $\checkmark$ &\centering $\times$ &\centering $\times$ & \centering $\times$ & \centering $\times$   &\centering $\checkmark$\tabularnewline	
\centering Learning Framework
&\centering Single-level &\centering Single-level &\centering Single-level &\centering Bilevel &\centering Bilevel  &\centering Bilevel\tabularnewline	
\centering Generalization Technique
& \centering $l_2$-ECN & \centering $\times$ & \centering $\times$ & \centering $\times$ & \centering $\times$   &\centering Uniform Stability \tabularnewline
\centering Generalization Bounds
& \centering $\mathcal{O}(m^{-1/2})$  & \centering $\times$ & \centering $\times$ & \centering $\times$ & \centering $\times$   &\centering $\mathcal{O}(T^c/m)$ \tabularnewline
\centering Regularization
&\centering $\ell_{2,1}$-norm  & \centering $\times$ & \centering $\times$ &\centering $\times$ &\centering $\times$   &\centering $\ell_{2,1}$-norm \tabularnewline	
\centering Loss Function Setting
& \centering Manual & \centering Manual  & \centering Manual & \centering Manual  & \centering Automatic   &\centering Automatic \tabularnewline	
\hline
\end{tabular}
}
\label{table1_highlight}
\end{table*}

\subsection{Related Work} 
The principle of meta-learning is to dynamically search for the best learning strategy to mitigate algorithmic bias \citep{vilalta2002perspective, shu2023learning}.
Following this strategy, Shu et al. \cite{shu2023learning} developed a meta-weighting network with a bilevel optimization framework to address corrupted and imbalanced data.
Moreover, the class information is added as the meta feature to learn a class-aware weighting policy \citep{shu2023cmw}, and a probabilistic formulation is proposed for meta-weight-net \citep{zhao2021probabilistic}. With the help of a small amount of unbiased meta data, these approaches realize adaptive weighting by additionally learning a mapping function from individual loss to corresponding weights. 
Simultaneously, Zhou et al. \citep{zhou2022core} also developed a probabilistic bilevel model for sample coreset selection, which could assign discrete 0/1 masks instead of continuous weights on each sample.
Different from the aforementioned approaches, the current work focuses on bilevel statistical learning models (i.e., sparse additive models) for robust estimation and variable selection.

To better highlight the novelty of our MAM, we summarize its properties in Table \ref{table1_highlight} compared with several related approaches, including some recent robust additive models, e.g., Sparse Additive Model (SpAM) \citep{liu2007spam}, Sparse Additive Quantile Regression Models (SAQR) \citep{lv2018oracle}, Sparse Modal Additive Model (SpMAM) \citep{chen2020sparse}, Correntropy-induced Sparse Additive Machine (CSAM) \citep{yuan2023sparse}, Global Importance Weighting (GIW) \citep{fang2024generalizing}, Kolmogorov-Arnold Network (KAN) \citep{liu2025kan}, Tilted Sparse Additive Model (TSpAM) \citep{wang2023tilted}, Concept-based Additive Taylor Model (CAT) \cite{duong2024cat}, Neural Additive Models (NAM) \cite{agarwal2021neural}, and the bilevel models like the Probabilistic Bilevel Coreset Selection (PBCS) \citep{zhou2022core} and Meta-Weight-Net (MWNet) \citep{shu2019meta}. Table \ref{table1_highlight} shows that the proposed MAM achieves auto weighting and sparse variable selection simultaneously.

\section{MAM: Meta Additive Models} \label{section2}

This section begins by recalling typical sparse additive models and then introduces the meta additive model (MAM) along with its computational algorithm.

\subsection{Sparse Additive Models}
Let $\ell(\cdot, \cdot): \mathbb{R} \times \mathbb{R}\rightarrow \mathbb{R}$ be a loss function, let  $\lambda>0$ be a regularization parameter, and define $\Omega(f)$ as a penalty for a hypothesis function $f\in\mathcal{H}$. 
To approximate the ground truth function $f^*$ in Eq.\eqref{data-model}, various additive models have been proposed under the principles of empirical risk minimization (ERM) \citep{stone1985additive} and structural risk minimization (SRM) \citep{kandasamy2016additive,liu2007spam,yuan2023sparse}, respectively.

In this paper, we form the additive hypothesis space based on the smoothing splines \citep{liu2007spam,wang2021huber}. 
Let $\left\{\psi_{j k}\right.$ : $k=1, \cdots, \infty\}$ be bounded and orthonormal basis functions on $\mathcal{X}_j$. Then the component function space can be defined as $\overline{\mathcal{H}}_j=\left\{\bar{f}_j: \bar{f}_j=\sum_{k=1}^{\infty} \beta_{j k} \psi_{j k}(\cdot),~~ \beta_{j k}\in\mathbb{R}\right\}, j=1, \cdots, p.$

After truncating these basis functions to finite dimension $d$, we get the hypothesis space of  component function 
\begin{equation} \label{component-space}
\mathcal{H}_j=\left\{f_j: f_j=\sum_{k=1}^d \beta_{j k} \psi_{j k}(\cdot)\right\}. 
\end{equation}

For each  input $x_i=$
$\left(x_{i 1}, \cdots, x_{i p}\right)^T \in \mathbb{R}^p$ from any given observations $\left\{\left(x_i, y_i\right)\right\}_{i=1}^n$, its  transformed version can be represented as $$\Psi_i=(\psi_{11}\left(x_{i 1}\right), \cdots, \psi_{1 d}\left(x_{i 1}\right), \cdots, \psi_{p d}(x_{i p}))^T\in\mathbb{R}^{p d}.$$

Denote $\boldsymbol{\beta}=\left(\beta_{11}, \cdots, \beta_{1 d}, \cdots, \beta_{p 1}, \cdots, \beta_{p d}\right)^T \in \mathbb{R}^{p d}$ as the objective coefficient. The estimated additive function can be represented as
\begin{equation} \label{additive_prediction_function}
\hat{f}=\sum_{j=1}^p \hat{f}_j~~ \mbox{with} ~~
\hat{f}_j
= \sum_{j=1}^p \sum_{k=1}^d \hat{\beta}_{j k} \psi_{j k}(\cdot)\in \mathcal{H}_j.
\end{equation}

Denote $\beta_j = (\beta_{j1},\cdots,\beta_{jd})\in \mathbb{R}^d$.
The weighted $\ell_{2,1}$-regularizer \citep{wang2023tilted,yuan2023sparse} is defined as
$$\Omega(f)=\sum_{j=1}^p \tau_j \sqrt{\left\|f_j\right\|_2^2}
=\sum_{j=1}^p \tau_j \sqrt{\sum_{k=1}^d \beta_{j,k}^2}  
=\sum_{j=1}^p \tau_j  \|\beta_{j}\|_2, \forall f=\sum_jf_j, f_j\in\mathcal{H}_j,$$
where positive weight $\tau_j$ is w.r.t. variable $X_j$, $j=1,...,p$.

After equipping the additive representation \eqref{additive_prediction_function} and the  weighted $\ell_{2,1}$-regularizer, we can formulate  single-level sparse additive models as
\begin{equation*}
\hat{\boldsymbol{\beta}}=\arg\min _{\boldsymbol{\beta} \in \mathbb{R}^{p d}}\left\{\frac{1}{n} \sum_{i=1}^n \ell \left(y_i,\Psi_i^{T} \boldsymbol{\beta}\right) + \lambda \sum_{j=1}^p \tau_j \|\beta_j\|_2\right\}.
\end{equation*}

Recently, some works have investigated robust losses $\ell$ under this framework, see e.g., 
the ramp loss \citep{chen2021sparse} and the correntropy-induced loss \citep{yuan2023sparse} for classification, and the mode-induced loss \citep{chen2020sparse} and the Huber loss \citep{wang2021huber} for regression. However, all the above approaches are limited to pre-specifying the loss. They are implemented using a single-level optimization scheme, which can rapidly deteriorate model performance under unfitting data conditions. 
Moreover, it is challenging to manually obtain the additional hyperparameters of robust loss functions \citep{shu2019meta, shu2023learning}. 

\begin{figure}[!t]
\centering
\includegraphics[width=\linewidth]{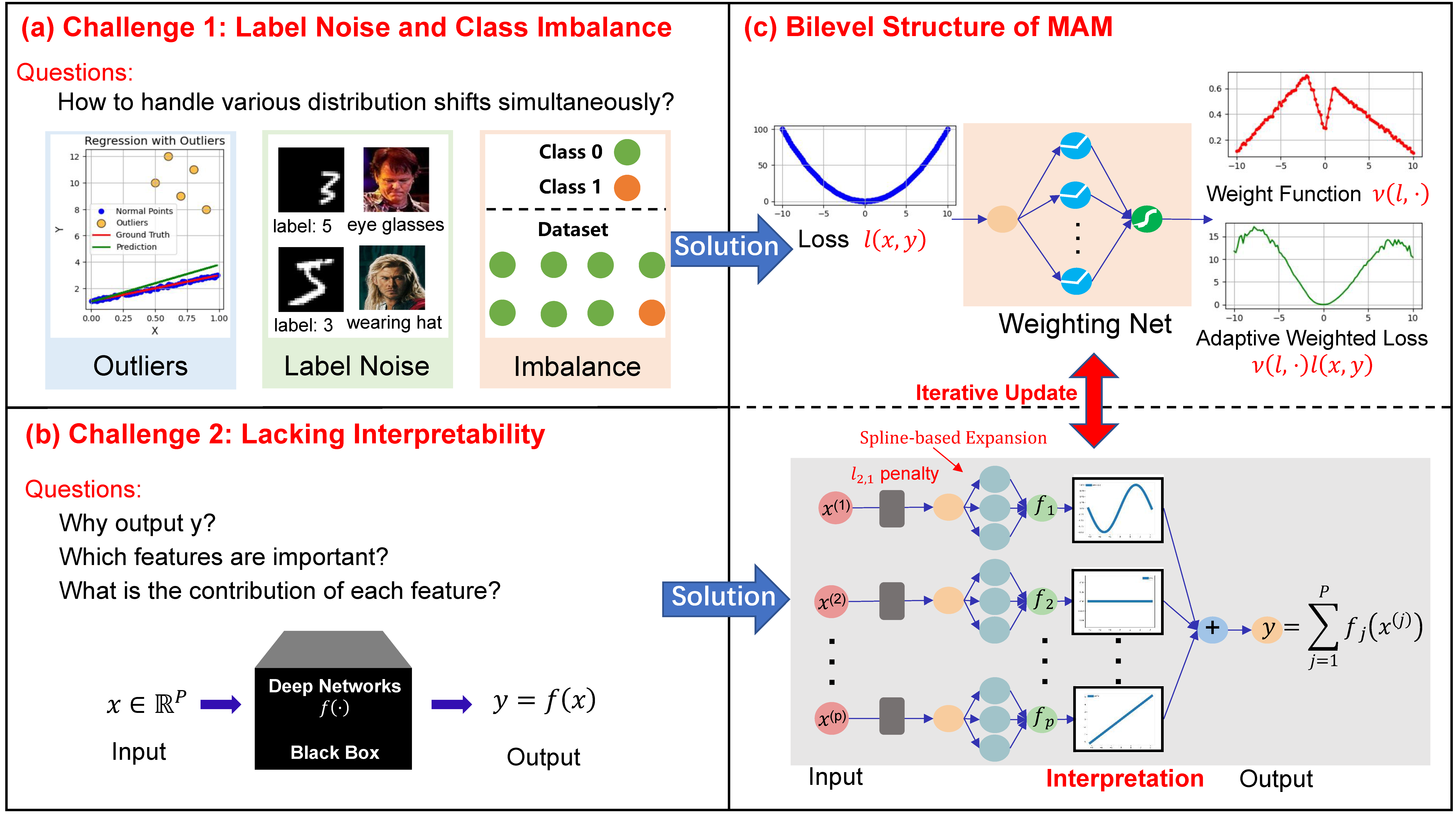}
\caption{The motivation and the bilevel structure of MAM}
\label{fig_bilevel_motivation}
\end{figure}

\subsection{Bilevel Additive Model}
As illustrated in Figure \ref{fig_bilevel_motivation}, we formulate a new additive model under bilevel optimization to achieve auto-weighting, sparse variable selection, and interpretable prediction, simultaneously. 
Denote $\mathcal{V}\left(L^{\text {train}}; \theta\right)$ as the meta weighting function whose input is the training loss $L^{\text {train}}$ and parameter is denoted as $\theta$. 
Let $\boldsymbol{\beta}$ be the parameter of the prediction function, which is updated in the lower level.

\noindent \textbf{Upper Level:}
Given a small meta (validation) dataset ${D}^{meta}=\{x_i,y_i\}_{i=1}^m$ (without outliers or noisy labels),  the objective function of the upper-level problem  can be described as follows,
\begin{equation} \label{upper_objective}
\begin{aligned}
\mathcal{L}^{meta}\left(\hat{\beta}(\theta) ; \theta\right)
=\frac{1}{m} \sum_{i=1}^{m} L_{i}^{meta}\left(\hat{\beta}(\theta)\right)
:=\frac{1}{m} \sum_{i=1}^{m} \ell (y_i, \Psi_i^T \hat{\beta}(\theta) ),
\end{aligned}
\end{equation}
where $\hat{\beta}(\cdot)$ is the parameter of the lower level objective function relevant to $\theta$, and $\theta$ is the parameter of the weighting network in the upper level. 

\noindent \textbf{Lower Level:}
Based on another $n$ training samples ${D}^{train}=\{x_i,y_i\}_{i=m+1}^{m+n}$, we can formulate  the lower level objective function  as
\begin{equation} \label{lower_objective}
\begin{aligned}
\mathcal{L}^{\text {train }}(\boldsymbol{\beta}; \theta)
=&\frac{1}{n} \sum_{i=m+1}^{m+n} \mathcal{V}\left(L_{i}^{\text {train }}(\hat{\boldsymbol{\beta}}) ; \theta\right) L_{i}^{\text {train }}(\boldsymbol{\beta}) + \lambda \sum_{j=1}^{p} \tau_j \|\beta_j\|_2 \\
:=&\frac{1}{n} \sum_{i=m+1}^{m+n} \mathcal{V}\left(\ell (y_i, \Psi_i^T \hat{\boldsymbol{\beta}} ); \theta\right) \ell (y_i, \Psi_i^T {\boldsymbol{\beta}}) + \lambda \sum_{j=1}^p \tau_j \sqrt{\sum_{k=1}^d \beta_{j,k}^2}.
\end{aligned}
\end{equation}

Here the objective parameter $\boldsymbol{\beta}=(\beta_1, \cdots, \beta_p)$, and $\beta_j = (\beta_{j1},\cdots,\beta_{jd})$ for $j=1,2,\cdots,p$, and $\mathcal{V}\left(L_{i}^{\text {train }}(\boldsymbol{\beta}) ; \theta\right)$ is the weighting function learned by the MLP with parameter $\theta$, which adaptively assigns weight to each individual loss. The weighting function is practically acquired through an MLP with the Sigmoid or Tanh activation function before the output, ensuring the boundedness within 1.
Similar with the existing bilevel framework \citep{shu2023cmw,zhou2022core}, we employ the squared loss  $\ell(y_i, \Psi_i^{T} \boldsymbol{\beta}) = (y_i - \Psi_i^{T} \boldsymbol{\beta})^2$ for additive regression tasks. 

Indeed, our MAM can also be applied to classification tasks, where the class-conditional probability can be estimated by 
$
{\rm{Prob}}(Y=1 \mid X=x)=\frac{\exp \left( \hat{f}(x_i) \right)}{1+\exp \left( \hat{f}(x_i) \right)}
$
with the predictive function $\hat{f}$ defined in \eqref{additive_prediction_function}.
The corresponding loss function can be formulated as
\begin{equation}\label{logisticloss}
\ell(y_i, \Psi_i^{T}\boldsymbol{ \beta}) =y_i \Psi_i^T \boldsymbol{\beta}-\log \left(1+e^{ \Psi_i^T\boldsymbol{\beta}}\right) .
\end{equation}

\subsection{Computing Algorithm}

\begin{figure}
\centering
\includegraphics[width=\linewidth]{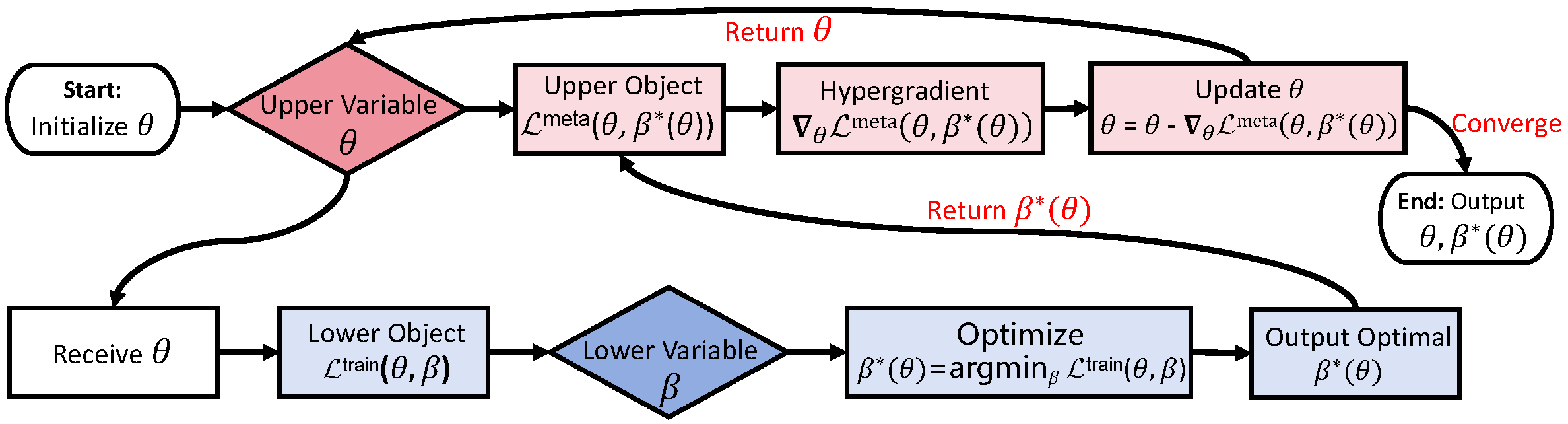}
\caption{Visualization of the bilevel optimization process}
\label{fig_bilevel_scheme}
\end{figure}

Denote $\eta_\beta^{(t)}$ and $\eta_\theta^{(t)}$ as the step sizes for updating upper parameter $\boldsymbol{\beta}$ and lower parameter $\theta$ in the $t$-th step, respectively.  
The MAM computing algorithm is a classical bilevel optimization \citep{franceschi2017forward}, as illustrated in Figure \ref{fig_bilevel_scheme} and summarized in Algorithm \ref{algorithm-1}. The optimization process of MAM mainly includes three steps as follows:

\begin{algorithm}[!b]
\caption{Computing algorithm of MAM}
\textbf{Input}: Meta data $D^{meta}$, training data $D^{train}$, batch size $b$, maximum iterations $T$, step sizes $\eta_\beta$, $\eta_\theta$.\\
\textbf{Initialization}: Initialize $\beta^{(0)}$, $\theta^{(0)}$.

\begin{algorithmic}[0] 
\FOR{$t=0$ to $T-1$}
{
\STATE 1) Randomly pick set:\\
~~~~~$(\hat{D}^{train})^{(t)} =$ Mini-Batch-Sampling$({D}^{train},b)$\\
~~~~~$(\hat{D}^{meta})^{(t)}  =$ Mini-Batch-Sampling$({D}^{meta},b)$\\
\STATE 2) Update $\hat{\beta}^{(t)}(\theta)$ based on Eq.\eqref{update_1} with $(\hat{D}^{train})^{(t)} $\\
\STATE 3) Update $\theta^{(t+1)}$ based on Eq.\eqref{update_2} with $(\hat{D}^{meta})^{(t)}$\\
\STATE 4) Update $\boldsymbol{\beta}^{(t+1)}$ based on Eq.\eqref{update_3} with $(\hat{D}^{train})^{(t)} $\\
}
\ENDFOR
\end{algorithmic}
\textbf{Output}: $\theta^{(T)}$ and $\hat{\beta}^{(T)}(\theta)$, $\hat{f}$ in Eq.\eqref{additive_prediction_function} with $\boldsymbol{\beta}^{(T)}$.
\label{algorithm-1}
\end{algorithm}

\textbf{Step 1. Update $\hat{\beta}(\theta) = (\hat{\beta}_1(\theta) ,\cdots,\hat{\beta}_p(\theta)) $:}
\begin{equation} \label{update_1} 
\begin{aligned}
\hat{\beta_j}^{(t)}(\theta) 
= &\beta_j^{(t)} - \frac{\eta_\beta^{(t)}}{n} \sum_{i=1}^n \left(\mathcal{V}\left(L_{i}^{\text {train }}(\mathbf{\boldsymbol{\beta}}) ; \theta\right) \nabla_{\beta_j} L_{i}^{\text {train}}(\mathbf{\boldsymbol{\beta}}) + \frac{\lambda \tau_j \beta_j}{\|\beta_j\|_2}\right) \bigg|_{\boldsymbol{\beta}^{(t)}}.
\end{aligned}
\end{equation}

\textbf{Step 2. Update $\theta$:}
\begin{equation}  \label{update_2}
\begin{aligned}
\theta^{(t+1)} = \theta^{(t)} - \frac{\eta_{\theta}^{(t)} \eta_\beta^{(t)}}{n} \sum_{i=1}^{n} (\frac{1}{m} \sum_{j=1}^m G_{i,j}) \frac{\partial \mathcal{V}\left(L_{i}^{\text {train }}(\mathbf{\beta}) ; \theta\right)}{\partial \theta} \bigg|_{{\theta}^{(t)}},
\end{aligned}
\end{equation}
where $G_{i,j} = \frac{\partial L_j^{meta}(\beta)}{\partial \beta}\bigg|_{\hat{\beta}^{(t)}(\theta)} \cdot \frac{\partial L_{i}^{\text {train}}(\mathbf{\beta})}{\partial {\beta}} \bigg|_{{\boldsymbol{\beta}}^{(t)}}$.

\textbf{Step 3. Update $\boldsymbol{\beta}=({\beta}_1,\cdots,{\beta}_p)$:}
\begin{equation}\label{update_3}
\begin{aligned}
{\beta_j}^{(t+1)}  = \beta_j^{(t)} - \frac{\eta_\beta^{(t)}}{n} \sum_{i=1}^n \bigg(\mathcal{V} & \left(L_{i}^{\text {train }}(\mathbf{\boldsymbol{\beta}}) ; \theta^{(t+1)}\right) \nabla_{\beta_j} L_{i}^{\text {train}}(\mathbf{\boldsymbol{\beta}}) 
+   \frac{\lambda \tau_j \beta_j}{\|\beta_j\|_2} \bigg) \bigg|_{\boldsymbol{\beta}^{(t)}}.
\end{aligned}
\end{equation}

\begin{remark}
This optimization process can be considered a combination of forward and reverse gradient computation for the bilevel learning problem \citep{shu2023learning}.
Due to the regularization in the additive space, the update process can be further decomposed into several steps with respect to each component function $ f_j = \beta_j \psi_j(\cdot)$ for $ j = 1, 2, \ldots, p$. 
\end{remark}

\section{Theoretical Analysis: Convergence, Consistency and Generalization} \label{section3}

\subsection{Analysis of Computing Convergence}

Before presenting our main results on algorithmic convergence, we recall some necessary assumptions involved in our analysis, which are commonly used in convergence analysis for bilevel optimization \citep{shu2023learning, shu2023cmw, DBLP:conf/nips/BaoWLZZ21}. 

\begin{assumption} \label{assumption_1}
The loss function $\ell$ is Lipschitz smooth with constant $L$, and has $\rho$-bounded gradients with respect to the training/meta dataset.
\end{assumption}

\begin{assumption}\label{assumption_2}
The weighting function $\mathcal{V}(\cdot)$ is differentiable with a $\delta$-bounded gradient and twice differentiable with its Hessian bounded by $\mathcal{B}$.
\end{assumption}

Based on the above assumptions, it can be further derived that the objective function of the upper problem $\mathcal{L}^{meta}\left(\mathbf{\beta}^{*}(\theta)\right)$ is also Lipschitz smooth w.r.t. $\theta$.

\begin{lemma}\citep{shu2019meta} \label{lemma1}
Let Assumptions 3 and 4 be true. Then the gradient of $\mathcal{L}^{meta}\left(\hat{\beta}^{(t)}(\theta);\theta\right)$ w.r.t. parameter $\theta$ in the $t$-th iteration is $L_{\mathcal{V}}$-Lipschitz continuous, where $L_{\mathcal{V}}=\eta_\beta\rho^{2}\left(\eta_\beta L \delta^{2}+\mathcal{B}\right)$.
\end{lemma}

For brevity, we denote $T$ as the total number of iterations.
\begin{theorem}\label{theorem1}
(Upper-level Problem) Suppose that Assumptions 3 and 4 hold. Let the step size $\eta_\beta^{(t)}$ satisfies $\eta_\beta^{(t)} \leq \min\{\frac{c_1}{t} ,\frac{1}{4L}\}$ for some $c_1>0$. Let the step size $\eta_\theta^{(t)} \leq \min\{ \frac{c_2}{\sqrt{t}}, \frac{1}{2L}, \frac{1}{2L_{\mathcal{V}}}\}$, where $c_2>0$, $\sum_{t=1}^{\infty} \eta_\theta^{(t)} \leq \infty$, and $\sum_{t=1}^{\infty} (\eta_\theta^{(t)})^{2} \leq \infty$. 
Then, we have
\begin{equation*}
\lim_{t \rightarrow \infty} \mathbb{E}\left[\left\|\nabla_\beta \mathcal{L}^{meta}\left(\hat{\beta}^{(t)}\left(\theta^{(t)}\right)\right)\right\|_{2}^{2}\right] = 0.
\end{equation*}
\end{theorem}

\begin{theorem} \label{theorem2}
(Lower-level Problem) Suppose that Assumptions 1-2 hold. Under the same settings of step sizes, let the penalty coefficients satisfy $\lambda \cdot \tau_j \leq \frac{c_3}{T}$ with $c_3>0$ for all $j \in \{1,\cdots,p\}$.
Then, we have
\begin{equation*}
\lim _{t \rightarrow \infty} \mathbb{E}\left[\left\|\nabla_\theta \mathcal{L}^{train}\left({\boldsymbol{\beta}}^{(t)} ; \theta^{(t+1)}\right)\right\|_{2}^{2}\right]=0.
\end{equation*}
\end{theorem}

\begin{remark}
Theorems \ref{theorem1} and \ref{theorem2} demonstrate that MAM converges to critical points of both the upper level (on meta data) and lower level (on training data) with convergence rates like $\mathcal{O}(1/\ln(T))$ under mild conditions, which extends previous results on MWNet \citep{shu2019meta,shu2023learning} to additive models under Tikhonv regularization scheme. As we know, the derived bilevel optimization analysis is the first step for additive models under a learning strategy. 
\end{remark}

\subsection{Analysis of Variable Selection Consistency}

Now, we establish the theoretical characteristic of variable selection consistency for the proposed MAM. The following assumptions employed here have been widely used in the theoretical analysis of additive models \citep{lv2018oracle,chen2020sparse,wang2023tilted}.

Here, we restate the bounded assumption on outputs and splines. Denote the transformation mapping in $j$-th dimension as $\psi_{j}(\cdot)$ from $\mathbb{R} \rightarrow \mathbb{R}^d$, e.g., $\psi_{j}(u) = (\psi_{j 1}(u), \cdots, \psi_{j d}(u))^T \in \mathbb{R}^d$.

\begin{assumption} \label{assumption_3}
The output $y\in[-S,S]$ with a positive constant $S$.
\end{assumption}

\begin{assumption} \label{assumption_4}
Transformed splines $\left\|\psi_{j}\right\|_{\infty}<\infty$ for $j\in[1,2,\cdots,p]$ and $\|f\|_{\infty}<\infty, \forall f \in \mathcal{H}$.
\end{assumption}

Theoretically, all variables can be divided into two disjoint groups: informative and irrelevant variables, respectively \citep{wang2017regularized}.

Denote 
$\mathcal{J}^*=\left\{1, \ldots, p^*\right\}, p^* \leq p,$ as the index set of truly informative variables. 
The $\ell_2$-norm of irrelevant variable coefficient $\beta_j$ for $j \notin \mathcal{J}^*$ is theoretically expected to satisfy $\left\|{\beta}_j \right\|_2=0$. Hence, define $\hat{\mathcal{J}}=\left\{j: {\|\hat{\beta}}_j\|_2 \neq 0\right\}$ as the set of estimated informative variables by the proposed MAM.

\begin{theorem} \label{theorem3}
Let Assumptions \ref{assumption_3} and \ref{assumption_4} be true. For MAM with the squared loss, if  $\tau_j \geq \frac{\sqrt{d}(\|f\|_{\infty}+S) \|\psi_j\|_\infty}{{\lambda}}$ for any $j>p^*$, there holds $\hat{\mathcal{J}} \subset \mathcal{J}^*$.
\end{theorem}

\begin{remark}
Theorem \ref{theorem3} illustrates that the nonlinear MAM with weighted squared loss can identify the truly informative variables by taking proper $\lambda$ and weight $\tau_j, j=1, \ldots, p$, which is consistent with previous analysis for sparse additive models \citep{liu2007spam,wang2023tilted}. 
As proved in Theorem \ref{theorem4}, similar guarantees still hold for MAM with logistic loss in Eq.\eqref{logisticloss} on the classification tasks. 
\end{remark}

\begin{theorem} \label{theorem4}
Under Assumption \ref{assumption_4}, for MAM with logistic loss, we have $\hat{\mathcal{J}} \subset \mathcal{J}^*$  if $\tau_j \geq 2 \frac{\sqrt{d} \|\psi_j\|_\infty}{\lambda }, \forall  j>p^*$.
\end{theorem}

\begin{remark}
For the binary classification task, Assumption \ref{assumption_3} is unnecessary due to the standard limitation for output where $y \in \{-1, 1\}$.
Indeed, Theorems \ref{theorem3} and \ref{theorem4} might still be able to be extended to other loss functions, e.g., the correntropy-induced loss \citep{yuan2023sparse}, the mode-induced loss \citep{chen2020sparse}, and the TERM framework \citep{wang2023tilted} under different parameter conditions.
\end{remark}

\subsection{Analysis of Algorithmic Generalization}

Theorems \ref{theorem1} and \ref{theorem2} state that this bilevel scheme exhibits promising algorithmic convergence, while leaving its generalization analysis open. Most existing techniques for generalization analysis, e.g., covering numbers and Rademacher complexity \citep{shi2013learning,lei2023generalization}, primarily focus on the capacity of the hypothesis space, overlooking the realistic details of optimization. Thus, in the following, we aim to derive an upper bound on the generalization gap using the algorithmic stability technique \citep{DBLP:conf/nips/BaoWLZZ21}, which is closely relevant to optimization strategies, e.g., step sizes and (stochastic) gradient descent, for bilevel problems.

Given distributions $\mathbb{D}_{meta}$, $\mathbb{D}_{train}$, we get the meta set 
\begin{equation*} 
D^{meta} := \{(x_i,y_i)\}_{i=1}^m = \{\xi_i\}_{i=1}^{m} \sim (\mathbb{D}_{meta})^{m},
\end{equation*}  and
the training set 
\begin{equation*} 
D^{train} := \{(x_i,y_i)\}_{i=m+1}^{m+n} = \{\zeta_i\}_{i=1}^{n} \sim  (\mathbb{D}_{train})^{n}
\end{equation*}  
by independent sampling, where $m$ and $n$ are the sample sizes. This paper focuses on the outer-level population risk w.r.t $\mathbb{D}_{meta}$ and empirical risk w.r.t $D^{meta}$ \footnote{Inspired from \citep{DBLP:conf/nips/BaoWLZZ21}, this paper considers adding corruptions to $D^{meta}$ to access the generalization behavior of the meta-learner \citep{thrun1998lifelong} at upper level.}, which are defined respectively as 
\begin{equation*} 
\begin{aligned}
R\left(\theta, \beta(\theta) \right)=\mathbb{E}_{\xi \sim \mathbb{D}_{meta}}[\ell(\theta, \beta(\theta) ; \xi)] 
~~ \text{and} ~~
\hat{R}_{D^{meta}}\left( \theta, \beta(\theta)  \right)=\frac{1}{m} \sum_{i=1}^{m}\left[  \ell\left(\theta, \beta(\theta)  ; \xi_i\right)\right],
\end{aligned}
\end{equation*}
where $\ell$ is an objective function rewritten from Eq.\eqref{upper_objective} and $\beta(\theta)$ is the inner model parameter given the upper parameter $\theta$.
Let $(\theta, \beta(\theta))$ be estimated by our stochastic Algorithm \ref{algorithm-1}, denoted by $A$, with data $D^{meta}, D^{train}$. That is 
\begin{equation*}
(\theta, \beta(\theta)) = A\left(D^{meta}, D^{train}\right).
\end{equation*}

Similar to the previous works \citep{DBLP:conf/nips/BaoWLZZ21,hoffer2017train}, to evaluate the approximated weighting policy, we define
\begin{equation}\label{generalization_gap}
\mathbb{E}_{A, D^{meta}, D^{train}}\left[R\left(A\left(D^{meta}, D^{train}\right)\right)-\hat{R}_{D^{meta}}\left(A\left(D^{meta}, D^{train}\right)\right)\right]
\end{equation}
in the upper (outer) level as the generalization gap of algorithm $A$, which measures the difference between the population risk $R(A)$ and the empirical risk $R_{D^{meta}}(A)$.

Before deriving the generalization bound, we first give the following Definition \ref{def_stability} of $\varepsilon$-uniform stability on validation in expectation.
\begin{definition} (Uniform Stability) \label{def_stability}
A randomized bilevel optimization algorithm $\mathbf{A}$ is $\varepsilon$-uniformly stable on meta set in expectation if for all meta (validation) datasets $D^{meta}, D^{meta \prime } \in (\mathbb{D}_{meta})^{m}$ such that $D^{meta}, D^{meta \prime }$ differ in at most one sample. Thus $\forall D^{train} \in (\mathbb{D}_{train})^{n}$ and $\xi \sim \mathbb{D}_{meta}$, we have
\begin{equation}
\mathbb{E}_{A}\left[\ell\left(\mathbf{A}\left(D^{meta}, D^{train}\right); \xi \right)-\ell\left(A\left(D^{meta \prime }, D^{train}\right); \xi\right)\right] \leq \varepsilon.
\end{equation}
\end{definition}

In the following, we also present some necessary assumptions, which have been commonly used for deriving the convergence rate \citep{ji2021bilevel,liang2023lower} or the generalization bounds \citep{DBLP:conf/nips/BaoWLZZ21} for bilevel optimization problems. Denote $\Theta$ and $\Gamma$ as the parameter space of $\theta$ and $\beta$. $\Xi$ is the data space for $\xi$.
\begin{assumption} \label{assumption_generalization}
The main assumptions are structured as follows:
\begin{itemize}
\item $\Theta$ and $\Gamma$ are compact and convex with non-empty interiors, and $\Xi$ is compact.
\item $\ell(\theta, \beta, \xi) \in C^{2}(\Omega)$, where $\Omega$ is an open set including $\Theta \times \Gamma \times \Xi$ (i.e., $\ell$ is second order continuously differentiable on $\Omega$ ).
\item $\ell(\theta, \beta, \xi) \in C^{3}(\Omega)$, where $\Omega$ is an open set including $\Theta \times \Gamma \times \Xi$ (i.e., $\ell$ is third order continuously differentiable on $\Omega$).
\end{itemize}
\end{assumption}

\begin{remark}
Indeed, another assumption on the smoothness of $\ell$ exists in \citep{DBLP:conf/nips/BaoWLZZ21}, which already holds in this paper under Assumptions \ref{assumption_1}-\ref{assumption_4}. The proof sketch is provided in Eq.\eqref{still_smooth}. Thus, we removed such a smoothness assumption.
\end{remark}

We then establish a connection between the defined uniform stability and the generalization gap in Theorem \ref{generalization_stability}, which follows from the above definition.

\begin{theorem}\label{generalization_stability}
If algorithm $A$ is $\varepsilon$-uniformly stable on meta set in expectation for $D^{meta} \in (\mathbb{D}_{meta})^m$ and $D^{train} \in (\mathbb{D}_{train})^n$, there holds 
\begin{equation*}
\left|\mathbb{E}_{A, D^{meta} , D^{train} } \left[R\left(A\left(D^{meta}, D^{train}\right)\right)-\hat{R}_{D^{meta}}\left(A\left(D^{meta}, D^{train}\right)\right)\right] \right| \leq \varepsilon.
\end{equation*}
\end{theorem}

Based on the Assumptions \ref{assumption_1}-\ref{generalization_stability} on upper and lower objective functions, Bao et al. \citep{DBLP:conf/nips/BaoWLZZ21} derived the conclusion that the upper objective function $\ell(\theta,\beta(\theta))$ also satisfies the continuity and smoothness properties w.r.t. $\theta$. The so-called inner loop in \citep{DBLP:conf/nips/BaoWLZZ21} is 1 in this paper. In practical settings, the weighting function is learned by an MLP, whose last layer is a Sigmoid or Tanh function. Thus $\|\mathcal{V}\|_\infty \leq 1$. For simplicity, the regularization term is ignored here. Based on the bounded Assumptions \ref{assumption_3} and \ref{assumption_4}, we know that $\|\ell\|_\infty$ is also bounded. Thus we can derive that the weighted lower loss function $\mathcal{V} \cdot \ell$ satisfies $L^\prime$-smoothness,
\begin{equation}\label{still_smooth}
L^\prime = 2  \rho \delta + L\|\mathcal{V}\|_\infty + B\|\ell\|_\infty = 2  \rho \delta + L + B\|\ell\|_\infty \leq \infty,
\end{equation}
where $\rho$, $L$, $\delta$ and $B$ are the smoothness and continuity coefficients from Assumptions \ref{assumption_1} and \ref{assumption_2}.

Inspired by \citep{DBLP:conf/nips/BaoWLZZ21}, we can also prove that the upper loss $\ell(\theta, \beta(\theta),\xi)$ is also smooth and continuous w.r.t. $\theta$ with specific coefficients.
\begin{lemma} (Theorem 3 in \citep{DBLP:conf/nips/BaoWLZZ21}).\label{lemma_upper_continuous_smooth}
Suppose Assumptions \ref{assumption_1}-\ref{assumption_generalization} holds. Then $\forall \xi \in \mathbb{D}_{meta}$, the upper objective function $\ell\left(\theta, \beta(\theta), \xi\right)$ as a function of $\theta$ is $\overline{L}=\mathcal{O}\left(1+\eta_\beta L^\prime\right)$ Lipschitz continuous and $\overline{\gamma}=\mathcal{O}\left(\left(1+\eta_\beta\left(L^\prime\right)\right)^{2}\right)$ smooth.
\end{lemma}

\begin{theorem}\label{Gen_bound}
Suppose Assumptions \ref{assumption_1}-\ref{assumption_generalization} hold.
Let $\kappa=\frac{c((1-1 / m) \overline{L})}{c((1-1 / m) \overline{L})+1}$ with the positive constant $c \leq \frac{s(\ell)}{2 (\overline{L})^2}$. 
Set the learning rate $\eta_\theta^{(t)} \leq \frac{c}{t}$. 
Then we have
\begin{equation}
\left|\mathbb{E} \left[R\left(A\left(D^{meta}, D^{train}\right)\right)-\hat{R}_{D^{meta}}\left(A\left(D^{meta}, D^{train}\right)\right)\right] \right| \leq \mathcal{O}\left( \frac{T^C}{m}\right),
\end{equation}
where the positive constant $C= \kappa <1$ is independent on $T$ and $m$.
\end{theorem}

\begin{remark}
Based on Theorem \ref{Gen_bound}, we can derive that the data size of the meta set $m$ ($\uparrow$). The training iterations $T$ ($\downarrow$) directly affect the generalization performance ($\uparrow$) of bilevel algorithms with proper learning rate (step sizes), which are closely relevant to our empirical practice. There are a few investigations \citep{DBLP:conf/nips/BaoWLZZ21} in the generalization analysis of bilevel optimization algorithms, where our result $\mathcal{O}\left( \frac{T^{C}}{m}\right)$ is comparable to their bounds, e.g., $\mathcal{O}\left( \frac{T^{C_1}}{m}\right)$ in \citep{DBLP:conf/nips/BaoWLZZ21}, or $\mathcal{O}\left( \frac{T^{C_2}}{m}\right)$ and $\mathcal{O}\left( \frac{T\ln(T)}{m}\right)$ in \citep{zhang2024sbo} without requirements on convexity, where $C<1$, $C_1<=1$ and $1<C_2<2$.
\end{remark}

\begin{remark}
The tightness of the derived generalization bound for MAM implies that the bilevel scheme significantly enhances the generalization capability, which also aligns with our motivation for equipping sparse additive models with a bilevel weighting policy \citep{shu2023learning} to learn from corrupted training data.
\end{remark}

\subsection{Complexity Analysis on Training Time and Storage Space}

As illustrated in Theorems \ref{theorem1} and \ref{theorem2}, the optimizers in the upper and lower problems reaches the convergence rates of $\mathcal{O}(\frac{\ln(T)}{\sqrt{T}})$ and $\mathcal{O}(\frac{1}{\sqrt{T}})$, respectively. 

The time and space complexity of Algorithm 1 for the Meta Additive Model (MAM) can be rigorously analyzed as follows. 

\begin{proposition} (Time Complexity)
Algorithm \ref{algorithm-1} has a time complexity of $\mathcal{O}(T \cdot (n+m) \cdot p \cdot d)$, where $T$ is the number of iterations, $n$ is the training set size, $m$ is the meta set size, $p$ is the number of variables, and $d$ is the number of basis functions per variable.
\end{proposition}

\begin{proposition} (Space Complexity)
Algorithm \ref{algorithm-1} has a space complexity of $\mathcal{O}((m+n) \cdot p + p \cdot d)$, where $m$ is the meta set size, $n$ is the training set size, $p$ is the number of variables, and $d$ is the number of basis functions per variable.
\end{proposition}

These complexity bounds demonstrate that MAM's computational efficiency scales linearly with problem dimensions, making it suitable for high-dimensional applications while maintaining theoretical tractability. Moreover, the empirical comparison of training time is shown in Figure \ref{complex_UCI}, where MAM achieves superior accuracy and robustness, albeit at a slightly higher computational cost.

\section{Experimental Evaluations}\label{section4}

This section validates the effectiveness of MAM on simulated and real-world data. All experiments are implemented in Python and MATLAB on a PC running Windows 10, an Intel i7 CPU, and an NVIDIA GeForce GTX 1660 Super GPU. 
These experiments are conducted on synthetic and real-world datasets under various corruptions in input features, labels, and category ratios to validate robust and imbalanced estimation, time cost, and feature selection.

\subsection{Experimental Setup}

\textbf{Baselines.}
For the regression tasks, we compare the proposed MAM with several variable selection algorithms, e.g., Lasso \citep{tib1994lasso}, SpAM \citep{liu2007spam}, Regularized Modal Regression (RMR) \citep{wang2017regularized} and TSpAM \citep{wang2023tilted}. Here, the squared loss is chosen as the loss function for SpAM and MAM. For simplicity, the TSpAM used here is accelerated by the random Fourier acceleration technique \citep{39}. Furthremore, we also compare with MLP, XGBoost, Correntropy-based Sparse Additive Machine (CSAM) \citep{yuan2023sparse}, Ramp Sparse Additive Model (RSAM) \cite{chen2021sparse}, Huber Additive Model (HAM) \cite{wang2021huber}, Global Importance Weighting (GIW) \cite{fang2024generalizing}, Kolmogorov–Arnold Networks (KAN) \cite{liu2025kan}, NAM \cite{agarwal2021neural}, Neural Basis Models (NBM) \cite{radenovic2022neural},  and Concept-based Taylor Additive Models (CAT) \cite{duong2024cat} with second-order concepts.
For the classification tasks, the competitors include SpAM \citep{liu2007spam}, SAM \citep{zhao2012sam}, linear $\ell_1$-SVM \citep{zhu2003norm} and TSpAM \citep{wang2023tilted}. The logistic loss is selected as the loss function for SpAM, TSpAM, and MAM for simplicity.
We repeat each experiment $50$ times and report the average results, along with the standard deviations, under the imbalanced categories and different types of label noises. 
Moreover, we analyze the corresponding time cost, present convergence curves, and validate the robustness of MAM against feature corruption on higher-dimensional UCI datasets in Section \ref{time_sec}. More real-world tabular and image datasets are also considered below. All used data is divided into a training set, a meta (validation) set, and a testing set with a ratio of $3:1:1$.

\textbf{Hyperparameter Selection.} For fairness, the penalty coefficient $\lambda$ is shared in the following experiments for all compared regularized methods, which is tuned across $\{10^{-6},10^{-5},10^{-4},10^{-3},10^{-2},10^{-1},1\}$. Let $\tau_j=1$ for all $j\in[1,2,\cdots,p]$, and the TERM-induced hyperparameters of TSpAM \citep{wang2023tilted} are tuned across $[\pm0.1,\pm0.5,\pm1,\pm2]$. The remaining baselines' hyperparameters are tuned according to their respective publications.

\textbf{Evaluation Criterion.}
For the regression tasks, the \emph{mean squared error (MSE)} is used to measure the difference between the ground truth $f^*(x)$ and the prediction $\hat{f}(x)$, i.e., $MSE=\frac{1}{n}\sum_{i=1}^n (\hat{f}(x_i)-y_i)^2$.  
Classification accuracy is used in classification scenarios. 
Inspired by \citep{wang2023tilted}, variable selection results are measured by the \emph{average selection percentage (ASP)}, which refers to the average probability of variables that are correctly identified. 
For completeness, the training time cost is also listed in the results, as shown in Figure \ref{complex_UCI}.

\begin{table*}[!t]
\centering
\caption{The ASP ($\uparrow$) and MSE (average MSE $\downarrow$ $\pm$ standard deviation $\downarrow$) on the synthetic corrupted data.}
\resizebox{\textwidth}{!}{
\begin{tabular}{cccccccccccccc}
\toprule 
\multirow{2}{*}{Noise Type} & \multicolumn{2}{c}{ Lasso} & \multicolumn{2}{c}{RMR} & \multicolumn{2}{c}{SpAM} & \multicolumn{2}{c}{TSpAM} & \multicolumn{2}{c}{MAM(ours)}\\
\cmidrule(r){2-3} \cmidrule(r){4-5} \cmidrule(r){6-7} \cmidrule(r){8-9}\cmidrule(r){10-11}
&  ASP  &   MSE
&  ASP  &   MSE
&  ASP  &   MSE
&  ASP  &   MSE
&  ASP  &   MSE \\
\midrule
$\epsilon^A$     &0.830     &  1.037 $\pm$ 0.119     &0.380 &0.597 $\pm$ 0.141
& 0.918 & 0.204 $\pm$ 0.049  & 0.925 &  0.157 $\pm$ 0.046 &  \textbf{0.931}     &   \textbf{0.081 $\pm$ 0.037} \\
$\epsilon^B$    &0.655     &  0.713 $\pm$ 0.057     &0.610 &0.425 $\pm$ 0.112   & 0.588 & 0.350 $\pm$ 0.078  & \textbf{1.000} & 0.019 $\pm$ 0.005 &  \textbf{1.000}&  \textbf{0.015 $\pm$ 0.005} \\
$\epsilon^C$     &0.865     & 1.546 $\pm$ 0.269    &0.843 & 0.247 $\pm$ 0.062   & 0.988 &  0.094 $\pm$ 0.060  & \textbf{1.000} &  0.037 $\pm$  {0.012} &  \textbf{{1.000}} &   \textbf{{0.029} $\pm$ 0.012}\\
\bottomrule
\end{tabular}}
\label{t2-simulation}
\end{table*}

\subsection{Experiments on Synthetic Data}

\noindent \textbf{Simulated regression task.}
Firstly, we generate $200$ samples with $8$ informative variables and $92$ irrelevant variables. Following \citep{liu2007spam,chen2020sparse}, the data for regression is generated from
\begin{equation} \label{simulated_regression}
Y=f^*(X)+\epsilon = \sum_{j=1}^{8} f_j(X_j)+\epsilon,
\end{equation}
where $f_1(u) = -2 \sin(2u), f_2(u)=8u^2, f_3(u)=\frac{7\sin u}{2-\sin u}, f_4(u)=6e^{-u}, f_5(u)=u^3+\frac{3}{2}(u-1)^2, f_6(u)=5u, f_7(u)=10\sin(e^{-u/2}), f_8(u)=-10\widetilde{\phi}(u,\frac{1}{2},\frac{4}{5}^2)$.
Notice that $\epsilon$ represents the noise and $\widetilde{\phi}$ stands for the normal cumulative distribution with mean of $\frac{1}{2}$ and the standard deviation of $\frac{4}{5}$. 
For the training data, three types of noises are considered here, including the noise $\epsilon^{A}$ following the skewed zero-mean distribution $\{0.8\mathcal{N}(-2,1)+0.2\mathcal{N}(8,1)\}$, the noise $\epsilon^B$ following the skewed zero-mode distribution $\{0.8\mathcal{N}(0,1)+0.2\mathcal{N}(20,1)\}$ and the noise $\epsilon^C$ following the heavy-tailed Student's $t$ distribution with freedom of 2. Note that the small meta set (validation set) for the proposed MAM is crucial for learning the proper weighting function, which is necessary to ensure unbiased results without corruption. To evaluate the prediction performance, the test data is also generated by the ground truth function \eqref{simulated_regression} with merely Gaussian noise $\epsilon \in \mathcal{N}(0,1)$. 

Table \ref{t2-simulation} shows that, under the zero-mean noise $\epsilon^A$, our proposed MAM, SpAM, and T-SpAM perform well in prediction and variable selection with nearly $ASE = 1$. However, the classical Lasso and RMR, which are limited to learning linear component functions, fail to fit the curve accurately.
In addition, especially under noise $\epsilon^B$ and $\epsilon^C$, the proposed MAM outperforms other competitors with the smallest MSE as well as the standard deviation. It can identify almost all truly informative variables.
The regression results verify the robustness of MAM against complex skewed noise.

\begin{table*}[!t]
\centering
\caption{The ASP ($\uparrow$) and accuracy (average ACC $\uparrow$ $\pm$ standard deviation $\downarrow$) on the synthetic corrupted data for classification. $r_1$ stands for the ratio of samples with noisy labels, and $r_2$ represents the imbalance factor.}
\resizebox{\textwidth}{!}{
\begin{tabular}{ccccccccccccccccccc}
\toprule 
\multirow{2}{*}{$r_1$} & \multirow{2}{*}{$r_2$}&\multicolumn{2}{c}{$\ell_1$-SVM} & \multicolumn{2}{c}{SAM} & \multicolumn{2}{c}{SpAM} & \multicolumn{2}{c}{TSpAM} & \multicolumn{2}{c}{MAM(ours)}\\
\cmidrule(r){3-4} \cmidrule(r){5-6} \cmidrule(r){7-8} \cmidrule(r){9-10}\cmidrule(r){11-12}
&   
&  ASP  &   ACC
&  ASP  &   ACC
&  ASP  &   ACC
&  ASP  &   ACC
&  ASP  &   ACC \\
\midrule
$10\%$   &$50\%$  &0.43       & 0.58 $\pm$ 0.04 &0.98 & 0.76$\pm$ 0.05  & 0.94& 0.79 $\pm$ 0.06 & 1.00 & 0.89 $\pm$ 0.03&  \textbf{1.00} &  \textbf{0.91 $\pm$ 0.02}\\
$30\%$   &$50\%$ &0.53       &  0.57 $\pm$ 0.05  &0.38 &  0.60 $\pm$ 0.04  & 0.41 &   0.63 $\pm$ 0.05 & \textbf{0.53}  & 0.70 $\pm$ 0.07 & \textbf{ 0.53 } & \textbf{0.72 $\pm$ 0.05}\\
$50\%$   &$50\%$ &0.23      &  0.53 $\pm$ 0.07 &0.03 & 0.50 $\pm$ 0.04   & 0.01 & 0.50 $\pm$ 0.05  & 0.27 & 0.50 $\pm$ 0.05 &  \textbf{0.31} &  \textbf{0.53 $\pm$ 0.04}\\
\bottomrule
\end{tabular}
}
\label{t3-simulation}
\end{table*}

\begin{table*}[!t]
\centering
\caption{The ASP ($\uparrow$) and accuracy (average ACC $\uparrow$ $\pm$ standard deviation $\downarrow$) on the synthetic imbalanced data.}
\resizebox{\textwidth}{!}{
\begin{tabular}{ccccccccccccccccccc}
\toprule 
\multirow{2}{*}{$r_1$} & \multirow{2}{*}{$r_2$}&\multicolumn{2}{c}{$\ell_1$-SVM} & \multicolumn{2}{c}{SAM} & \multicolumn{2}{c}{SpAM} & \multicolumn{2}{c}{TSpAM} & \multicolumn{2}{c}{MAM(ours)}\\
\cmidrule(r){3-4} \cmidrule(r){5-6} \cmidrule(r){7-8} \cmidrule(r){9-10}\cmidrule(r){11-12}
&   
&  ASP  &   ACC
&  ASP  &   ACC
&  ASP  &   ACC
&  ASP  &   ACC
&  ASP  &   ACC \\
\midrule
$0$ & $5\%$      &1.00        &  0.50 $\pm$ 0.00 &1.00 & 0.80 $\pm$ 0.05  &1.00& 0.72 $\pm$ 0.10 & 1.00 & 0.83 $\pm$ 0.04 & 1.00 & \textbf{0.87 $\pm$ 0.04} \\
$0$ &$10\%$    &1.00        & 0.50 $\pm$ 0.00 &1.00 &0.86 $\pm$ 0.04  & 1.00& 0.82 $\pm$ 0.06 & 1.00& 0.88 $\pm$ 0.03 & 1.00& \textbf{0.91 $\pm$ 0.03}\\
$0$ & $15\%$    &1.00       &  0.50 $\pm$ 0.00 &1.00& 0.89 $\pm$ 0.03  & 1.00 & 0.86 $\pm$ 0.05 & 1.00 & 0.91 $\pm$ 0.03 & 1.00 & \textbf{0.92 $\pm$ 0.03}\\
\bottomrule
\end{tabular}
}
\label{t4-simulation}
\end{table*}

\noindent \textbf{Simulated classification task:}
We now evaluate the effectiveness of MAM on classification tasks with noisy labels, imbalanced categories, or both. 

\textbf{1) Robust Classification:}
Firstly, the robustness of MAM to noisy labels is assessed.
Similar with the experimental design in \citep{chen2020sparse, wang2023tilted}, the additive discriminant function is  formulated as
\begin{equation}\label{classfication_model}
f^*(x_i) = (x_{i1}-0.5)^2 + (x_{i2}-0.5)^2 -0.08,
\end{equation}
where $x_{ij}=(W_{ij}+U_i)/2$. $W_{ij}$ and $U_i$ are independently from $U(0,1)$ for $i=1, \cdots, 200$, $j=1, \cdots, 100$. 
The label satisfies $y_i=0$ when $f(x_i) \le 0$ and $1$ otherwise.

Consider the training dataset includes noisy labels at a ratio of $r_1$. The corruption rate $r_1$ is consistent across the two types of imbalanced samples.
Table \ref{t3-simulation} presents the classification accuracy and ASP of variable selection under different noisy labels, indicating that MAM exhibits competitive or even superior robustness compared to other baselines.

\textbf{2) Imbalanced Classification:}
We verify MAM's performance in handling imbalanced classification problems. The training data is generated from Eq.\eqref{classfication_model} with 
dimension $p=10$ here following the settings in \citep{wang2023tilted} instead of $p=100$. Denote $r_2$ as the imbalance factor, which represents the ratio of the negative class ($y_i = 0$) in the whole training dataset. Experimental results in 
Table \ref{t4-simulation} verifies that MAM achieves higher accuracy and a smaller standard deviation than the other approaches.

\begin{table*}[!t]
\centering
\caption{The average accuracy $\uparrow$ $\pm$ standard deviation $\downarrow$ on the synthetic corrupted and imbalanced data.}
\resizebox{\textwidth}{!}{
\begin{tabular}{ccccccccccccccccccc}
\toprule 
{$r_1$ } & {$r_2$}&{$\ell_1$-SVM} & {SAM} & {SpAM} &{TSpAM} &{MAM(ours)}\\
\midrule
$10\%$ & $15\%$      &0.50 $\pm$ 0.01 & 0.80 $\pm$ 0.05 & 0.79 $\pm$ 0.06 & 0.83 $\pm$ 0.04 & \textbf{0.86 $\pm$ 0.01 }\\
$10\%$ & $10\%$      & 0.50 $\pm$ 0.01 &0.76 $\pm$ 0.07 &0.71 $\pm$ 0.08 &  0.78 $\pm$ 0.04 &  \textbf{0.82 $\pm$ 0.03 }\\
$30\%$ & $15\%$      &  0.50 $\pm$ 0.02 &0.64 $\pm$ 0.05  &  0.68 $\pm$ 0.09 &  0.68 $\pm$ 0.07 &  \textbf{0.71 $\pm$ 0.05}\\
$30\%$ & $10\%$      & 0.50 $\pm$ 0.01 &0.60 $\pm$ 0.07 &0.62 $\pm$ 0.11  & 0.66 $\pm$ 0.08 & \textbf{0.70 $\pm$ 0.05}\\
\bottomrule
\end{tabular}
}
\label{t5-simulation}
\end{table*}

\textbf{3) Multi-objective Classification:}
Now we evaluate MAM's capability to tackle both robust and imbalanced classification tasks simultaneously.
For completeness, we consider several combinations of corrupted ratio $r_1\in\{0.1,0.3\}$ and imbalanced factor $r_2\in\{0.10,0.15\}$ with the same data-generation model \eqref{classfication_model}, where the sample size $n = 2000$ and the dimension $p = 10$. The imbalanced factor $r_2$ represents the ratio of the sample size in class 1 to the sample size in class 0.
Table \ref{t5-simulation} implies that MAM slightly outperforms the other baselines. Although TSpAM sometimes performs similarly, MAM is more flexible, as it does not require manual designation of a loss function or relevant loss hyperparameters for tuning.

\subsection{Experiments on UCI dataset}\label{time_sec}

Here, we evaluate the performance of MAM on the UCI datasets \footnote{Downloaded from \url{http://archive.ics.uci.edu/ml}.} for classification. The exploited datasets consist of the Balance Scale (Balance), Haberman’s Survival (Haberman), Statlog of Heart (Statlog), Spect, and Breast Cancer (Breast), which have been widely employed to evaluate the prediction performance (of additive models) for classification tasks \citep{lahiri2016forward}. The total instances of each dataset are split randomly into training, validation (meta), and test sets with a ratio of 3:1:1. 

\textbf{Performance on real-world data.} Table \ref{t6-uci} reports the average classification accuracy and standard deviation after 20 repeats, which further verifies the competitiveness of MAM in real applications. The result of  CSAM \citep{yuan2023sparse} with $\ell_{2,1}$ penalty is also reported, where the relevant parameter is tuned across $[0.5,1,1.5,\cdots,10]$.

\begin{table*}[!t]
\centering
\caption{The average accuracy $\uparrow$ and standard deviation $\downarrow$ on the UCI dataset for classification}
\resizebox{\textwidth}{!}{
\begin{tabular}{l|cccccccccccccccccc}
\toprule 
{dataset} &{$\ell_1$-SVM} & {SAM}  & {CSAM} &{TSpAM} &{MAM(ours)}\\
\midrule
Balance    & 78.45$\pm$2.39 &96.06$\pm$1.11  & 97.23$\pm$1.58 &  97.95$\pm$1.71 &  \textbf{98.11$\pm$1.27}\\
Haberman     & 72.55$\pm$3.83 &71.96$\pm$4.17   & 74.62$\pm$3.20 & 74.85$\pm$2.81& \textbf{75.37$\pm$2.19} \\
Statlog    & 79.62$\pm$7.03  &82.10$\pm$4.00  & 84.69$\pm$3.16 & 86.11$\pm$2.99 & \textbf{86.26$\pm$2.04}\\
Spect     & 80.54$\pm$4.27 &78.77$\pm$3.66   & 80.99$\pm$3.70 & 81.55$\pm$3.30 & \textbf{83.13$\pm$2.11} \\
Breast    & 95.54$\pm$0.70  &95.78$\pm$1.31   & 97.29$\pm$1.15 & 97.10$\pm$1.22 & \textbf{97.85$\pm$0.95}\\
\bottomrule
\end{tabular}
}
\label{t6-uci}
\end{table*}

\textbf{Performance on corrupted real-world data.} To further validate the training time and robustness of MAM, we conduct additional experiments on other UCI datasets. Table \ref{complex_UCI} records the time cost and predictive accuracy under both label and feature corruptions. 
Besides the label noises (with a sample ratio of $r_1$), we also generate noises following the Student (T) distribution with freedom of 2. Then the noises are added into the input $X$ (with a sample ratio of $r_3$) instead of the label $y$ to evaluate the robustness of MAM and some baselines against noisy features.
We add two robust meta-based models as baselines for fairness and comprehensiveness, namely MWNet \citep{shu2019meta} and probabilistic bilevel coreset selection (PBCS) \citep{zhou2022core}. The early-stop threshold is set to $10^{-6}$, and the time cost involves the hyperparameter selection process, which is independently repeated 10 times. $r_1$ and $r_3$ represent the ratios of samples with noisy labels or noisy features, respectively.
As shown in Table \ref{complex_UCI}, MAM achieves the highest prediction accuracy and the smallest standard deviation among the robust TSpAM. Random feature corruptions obviously degrade the performance of $\ell_1$ SVM, further validating the superiority and robustness of MAM against both label and feature corruptions. The time cost of MAM is higher than $\ell_1$-SVM due to additional automatic parameter searching. 

\begin{table*}[!t]
\centering
\caption{The training time (minute) ($\downarrow$) and average accuracy ($\uparrow$) $\pm$ standard deviation ($\downarrow$) on the new UCI data with different sample ratios of label corruption ($r_1$) and feature corruption ($r_3$) for classification.}
\resizebox{\textwidth}{!}{
\begin{tabular}{ccccccccccccccccccc}
\toprule 
\multirow{2}{*}{Dataset} & \multirow{2}{*}{$r_1 / r_3$}&\multicolumn{2}{c}{$\ell_1$-SVM} & \multicolumn{2}{c}{TSpAM} & \multicolumn{2}{c}{MWNet} & \multicolumn{2}{c}{PBCS} & \multicolumn{2}{c}{MAM(ours)}\\
\cmidrule(r){3-4} \cmidrule(r){5-6} \cmidrule(r){7-8} \cmidrule(r){9-10}\cmidrule(r){11-12}
&   
&  Time  &   ACC
&  Time  &   ACC
&  Time  &   ACC
&  Time  &   ACC
&  Time  &   ACC \\ 
\midrule
Z-Alizadeh   &$0/0$  &\textbf{0.41}       & 0.72$\pm$0.01 &6.78 & 0.73$\pm$0.05  & 6.71& \textbf{0.76$\pm$0.09} & 7.23 & 0.74$\pm$0.11&  6.81&  \textbf{0.76$\pm$0.06}\\
Z-Alizadeh   &$0.1/0$ &\textbf{0.43}       &  0.65$\pm$0.02  &6.83 &  0.68$\pm$0.10  & 6.89 &   0.70$\pm$0.14 & 7.45  & 0.70$\pm$0.14 & 7.01 & \textbf{0.74$\pm$0.05}\\
Z-Alizadeh    &$0/0.1$ &\textbf{0.43}      &  0.67$\pm$0.01 &6.84 & 0.70$\pm$0.09   & 6.81 & 0.71$\pm$0.12  & 7.25 & 0.72$\pm$0.11 &  6.93 &  \textbf{0.75$\pm$0.07}\\
Z-Alizadeh   &$0.1/0.1$  &\textbf{0.46}       & 0.61$\pm$0.02 &6.94 & 0.66$\pm$0.13  & 7.15& 0.69$\pm$0.14 & 7.95 & 0.68$\pm$0.16& 7.07 & \textbf{0.72$\pm$0.09}\\
Fertility   &$0/0$ &\textbf{0.21}       &  0.87$\pm$0.03  &4.82 &  0.87$\pm$0.04  & 4.11 &   0.89$\pm$0.05 & 4.83 & 0.88$\pm$0.04 & 4.36  & \textbf{0.90$\pm$0.02}\\
Fertility    &$0.1/0$ &\textbf{0.22}      &  0.81$\pm$0.05 &4.95 & 0.82$\pm$0.07   & 4.37 & 0.84$\pm$0.09  & 5.05 & 0.86$\pm$0.05 &  4.85 &  \textbf{0.88$\pm$0.04}\\
Fertility   &$0/0.1$ &\textbf{0.22}       &  0.83$\pm$0.05  &4.93 &  0.83$\pm$0.06  & 4.25 &   0.85$\pm$0.09 & 4.96 & 0.85$\pm$0.07 & 4.75 & \textbf{0.88$\pm$0.05}\\
Fertility    &$0.1/0.1$ &\textbf{0.22}     &  0.77$\pm$0.07 &5.42 & 0.79$\pm$0.11   & 4.56 & 0.82$\pm$0.13  & 5.37 & 0.81$\pm$0.14 &  5.02 &  \textbf{0.86$\pm$0.08}\\
\bottomrule
\end{tabular}}
\label{complex_UCI}
\end{table*}

\begin{figure}[!t]
\centering
\includegraphics[width=0.55\textwidth]{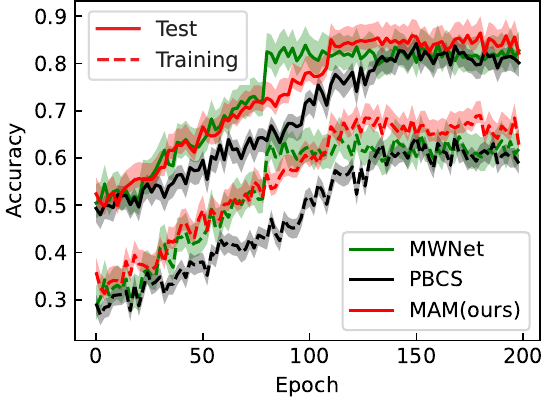}
\caption{Convergence curves of prediction accuracy versus training epoch for meta-based methods on corrupted Fertility dataset with 10\% noisy features and 10\% noisy labels.}
\label{fig:convergence}
\end{figure}

Indeed, the time cost of our spline-based MAM is comparable to kernel-based TSpAM and other meta-based models (MWNet \citep{shu2019meta} and PBCS \citep{zhou2022core}), which is further validated by the convergence curve in Figure \ref{fig:convergence}.

\subsection{Applications on Real-world Tabular Data}\label{sec_cme}

The collected tabular datasets include Airbnb listings with $n=275,598$ instances and $p=126$ dimensions, Coronal Mass Ejection (CME) datasets \cite{liu2018new} ($n=137$, $p=21$), and Alzheimer's Disease Neuroimaging Initiative (ADNI) clinical records (see adni.loni.usc.edu) with $n=795$ instances and $p=326$ dimensions.
Furthermore, the training, validation, and testing sets are sampled at the ratio of 8:1:1. Following \cite{bao2024robust}, a percentage of "$r_1$" samples from the training set are corrupted by adding noise $\epsilon \in \mathcal{N}(100,100)$ for regression.

\begin{table*}
\caption{Experiments with average MSE ($\downarrow$) $\pm$ standard deviation for regression. The upper, middle and lower panels stand for results on Airbnb listings, CME and ADNI clinical tabular datasets. Notably, $r_1$ stands for the rate of noisy labels (outliers). Besides, the 1st and 2nd best results among all baselines are highlighted in \textbf{bold} or \underline{underlined}, respectively.}
\label{regression}
\centering
\resizebox{\textwidth}{!}{
\begin{tabular}{c|c|c|c|c|c|c|c|c|c}
\hline
$r_1$ & MLP & XGBoost & TSpAM & HAM & KAN & NAM & NBM &CAT &  MAM\\
\hline
0 & 0.395 $\pm$ 0.093 & 0.363 $\pm$ 0.101& 0.421 $\pm$ 0.071& 0.414 $\pm$ 0.075& 0.423 $\pm$ 0.120&  0.446 $\pm$ 0.065 & 0.441 $\pm$ 0.040 & \underline{0.301 $\pm$ 0.034}  & \textbf{0.294 $\pm$ 0.027}\\

10\% & 0.517 $\pm$ 0.183 & 0.488 $\pm$ 0.191& 0.451 $\pm$ 0.094& \underline{0.448 $\pm$ 0.093} & 0.518 $\pm$ 0.202&  0.511 $\pm$ 0.180 & 0.504 $\pm$ 0.154 & 0.471 $\pm$ 0.106  & \textbf{0.328 $\pm$ 0.039}\\

30\% & 1.337 $\pm$ 0.422 & 1.391 $\pm$ 0.407& 0.513 $\pm$ 0.122& \underline{0.508 $\pm$ 0.109} & 1.413 $\pm$ 1.231 & 1.391 $\pm$ 1.211 & 1.386 $\pm$ 1.177 & 1.074 $\pm$ 0.992  & \textbf{0.362 $\pm$ 0.086}\\
\hline
0 & 0.598 $\pm$ 0.129 & 0.604 $\pm$ 0.140& 0.613 $\pm$ 0.164& 0.626 $\pm$ 0.144& 0.608 $\pm$ 0.103&  0.619 $\pm$ 0.116 & 0.613 $\pm$ 0.098 & \underline{0.585 $\pm$ 0.107}  & \textbf{0.568 $\pm$ 0.084}\\

10\% & 0.863 $\pm$ 0.217 & 0.844 $\pm$ 0.194& \underline{0.690 $\pm$ 0.181}& 0.705 $\pm$ 0.173& 0.858 $\pm$ 0.186&  0.874 $\pm$ 0.199 & 0.867 $\pm$ 0.159 & 0.830 $\pm$ 0.155  & \textbf{0.594 $\pm$ 0.085}\\

30\% & 1.611 $\pm$ 0.743 & 1.581 $\pm$ 0.726& 0.748 $\pm$ 0.206  & \underline{0.740 $\pm$ 0.183} & 1.577 $\pm$ 0.512  &  1.603 $\pm$ 0.481 & 1.584 $\pm$ 0.459 & 1.310 $\pm$ 0.422   & \textbf{0.656 $\pm$ 0.128}\\
\hline
0 & 0.174 $\pm$ 0.042 & \textbf{0.167 $\pm$ 0.037}& 0.175 $\pm$ 0.044& \underline{0.168 $\pm$ 0.040} & 0.184 $\pm$ 0.051&  0.190 $\pm$ 0.048 & 0.182 $\pm$ 0.041 & 0.175 $\pm$ 0.045  & 0.176 $\pm$ 0.039\\
10\% & 0.342 $\pm$ 0.091 & 0.336 $\pm$ 0.094& 0.194 $\pm$ 0.067& \underline{0.192 $\pm$ 0.059} & 0.361 $\pm$ 0.120 &  0.372 $\pm$ 0.114 & 0.364 $\pm$ 0.108 & 0.324 $\pm$ 0.092  & \textbf{0.187 $\pm$ 0.057}\\
30\% & 0.746 $\pm$ 0.139 & 0.739 $\pm$ 0.128& 0.247 $\pm$ 0.093 & \underline{0.251 $\pm$ 0.088} & 0.701 $\pm$ 0.122 &  0.713 $\pm$ 0.132 & 0.706 $\pm$ 0.119 & 0.649 $\pm$ 0.102  & \textbf{0.228 $\pm$ 0.080}\\
\hline
\end{tabular}}
\end{table*}

Table \ref{regression} presents results after 20 repeats with $r_1 \in [0, 10\%, 30\%]$ noisy training samples, showing that existing additive models are usually sensitive to label noise. Compared to baselines with robust loss (e.g., HAM), the proposed MAM also provides minor errors and better stability.
Without manual selection on additional hyperparameters, our MAM realizes the smallest MSE ($0.60\pm 0.14$).

\subsection{Applications on Real-world Image Data}

This subsection presents the results on two image datasets, e.g., the MNIST digits \cite{lecun1998mnist} ($n=70,000$ and $p=28\times28$) and CelebAMask-HQ images \cite{lee2020maskgan} ($n=30,000$ and $p=512\times512$). 
Inspired by \cite{duong2024cat}, the disentangled representation technique is used to extract 6 and 9 high-level factors from high-dimensional MNIST and CelebAMask-HQ datasets for training. Similar to the robust regression tasks in Section \ref{sec_cme}, the ratios of $r_1$ corrupted samples with noisy labels are generated by flipping their labels to the next (or first) category for classification. The maximum ratio between the sizes of major and minor classes is denoted as "$r_2$".  

\begin{table*}
\caption{Experiments with average Macro-F1 score ($\uparrow$) $\pm$ standard deviation for classification. The upper and lower panels stand for results on MNIST and CelebA images. Notably, $r_1$ and $r_2$ represent the rates of noisy labels and category imbalance, respectively.}
\label{classification}
\centering
\resizebox{\textwidth}{!}{
\begin{tabular}{c|c|c|c|c|c|c|c|c|c|c|c}
\hline
$r_1$& $r_2$& MLP & XGBoost & SpAM & CSAM & RSAM & GIW & KAN & NAM &CAT & MAM \\
\hline
0 & 1:1 & 0.942 $\pm$ 0.031 & 0.941 $\pm$ 0.040& 0.913 $\pm$ 0.061& 0.921 $\pm$ 0.057& 0.917 $\pm$ 0.077&  0.933 $\pm$ 0.034 & 0.901 $\pm$ 0.055 & 0.864 $\pm$ 0.046 & \textbf{0.949 $\pm$ 0.031} & \underline{0.945 $\pm$ 0.034}\\

0 & 1:10 & 0.891 $\pm$ 0.068 & 0.894 $\pm$ 0.067& 0.856 $\pm$ 0.089& 0.864 $\pm$ 0.071& 0.860 $\pm$ 0.083&  \underline{0.912 $\pm$ 0.051} & 0.853 $\pm$ 0.096 & 0.831 $\pm$ 0.109 & 0.898 $\pm$ 0.053 & \textbf{0.914 $\pm$ 0.046}\\

20\% & 1:10 & 0.830 $\pm$ 0.124 & 0.834 $\pm$ 0.117& 0.811 $\pm$ 0.130& 0.849 $\pm$ 0.088& 0.851 $\pm$ 0.091&  \underline{0.903 $\pm$ 0.074} & 0.820 $\pm$ 0.110 & 0.804 $\pm$ 0.166 & 0.839 $\pm$ 0.117 & \textbf{0.905 $\pm$ 0.059}\\
\hline

0 & 1:1 & \underline{0.766 $\pm$ 0.071} & 0.754  $\pm$ 0.076& 0.715 $\pm$ 0.089& 0.721 $\pm$ 0.082& 0.723 $\pm$ 0.082&  0.730 $\pm$ 0.101 & 0.746 $\pm$ 0.098 & 0.723 $\pm$ 0.113 & 0.740 $\pm$ 0.087 & \textbf{0.770 $\pm$ 0.060}\\

0 & 1:10 & 0.673 $\pm$ 0.106 & 0.661 $\pm$ 0.114& 0.634 $\pm$ 0.130& 0.637 $\pm$ 0.122& 0.644 $\pm$ 0.110&  \underline{0.725 $\pm$ 0.127} & 0.671 $\pm$ 0.124 & 0.630 $\pm$ 0.135 & 0.651 $\pm$ 0.104 & \textbf{0.759$\pm$ 0.069}\\

20\% & 1:10 & 0.611 $\pm$ 0.137 & 0.594 $\pm$ 0.142& 0.598 $\pm$ 0.151 & 0.620 $\pm$ 0.133& 0.617 $\pm$ 0.130&  \underline{0.712 $\pm$ 0.142} & 0.614 $\pm$ 0.148 & 0.599 $\pm$ 0.157 & 0.618 $\pm$ 0.121 & \textbf{0.736 $\pm$ 0.073}\\
\hline
\end{tabular}}
\end{table*}

In Table \ref{classification}, both the rates of noisy label ($r_1$) and imbalance factor ($r_2$) are considered. The case of $r_1$=0\% and $r_2$=$1:10$ shows the vulnerability of (neural) additive models to class imbalance. The results under $r_1$=20\% and $r_2$=$1:10$ imply that those carefully designed robust losses do not apply to the imbalance case. In contrast, the data-driven weighted loss of MAM performs well across both complex scenarios. Besides, MAM does not require manual selection of hyperparameters of the loss compared to robust CSAM and RSAM.

\section{Extended Analysis on Algorithmic Properties}

\subsection{Empirical Curves of Weighted Losses}

\begin{figure*}[!t]
\centering
\subfigure[Regression losses]{
\includegraphics[width=0.4\textwidth]{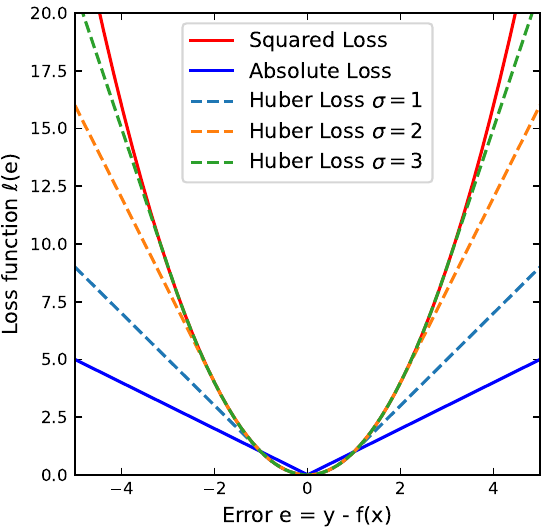}
}
\hspace{3mm}
\subfigure[MAM with squared loss]{
\includegraphics[width=0.383\textwidth]{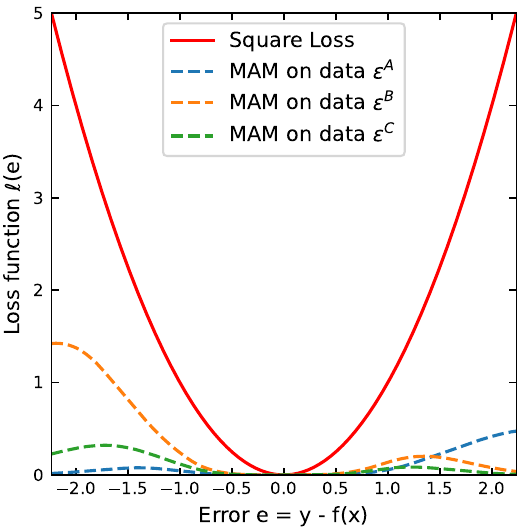}
}
\subfigure[Classification losses]{
\includegraphics[width=0.405\textwidth]{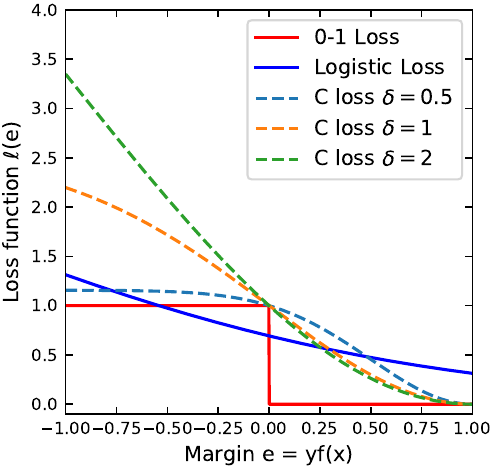}
}
\subfigure[MAM with logistic loss]{
\includegraphics[width=0.41\textwidth]{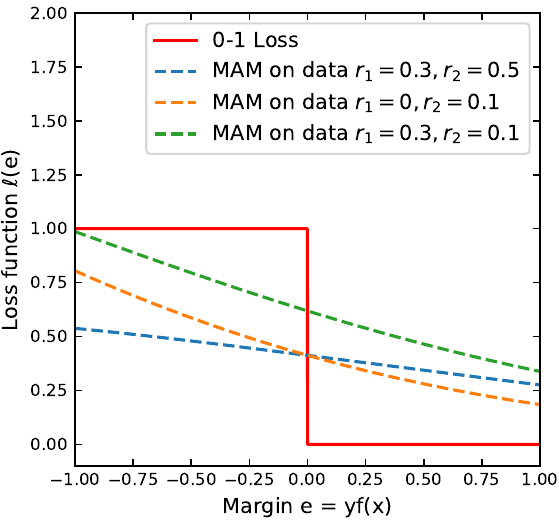}
}
\caption{The weighting curves of several loss functions and MAM. (a) The properties and curves of typical robust loss functions are dependent on the manual selection of specific hyperparameters. And their curve patterns are usually symmetric and fixed. 
(b) MAM can automatically learn the data-driven weighting function under different scenarios without additional hyperparameters. (c) and (d) are the relevant results for classification. These MAM curves are examples of weighted loss.}
\label{f1_weighting_curve}
\end{figure*}

In Figure \ref{f1_weighting_curve}, we investigate the implied weighting policy of MAM and several robust competitors for regression and classification tasks. Obviously, the curve patterns of these robust losses, including Huber loss \citep{wang2021huber}, and the correntropy-induced loss \citep{hu2013learning,fan2016consistency,yuan2023sparse}, are mainly fixed w.r.t. their hyperparameters. 
However, the MAM weighting function is automatically learned from the meta set, without additional hyperparameters to be manually selected. The weighting function of MAM can theoretically approximate any continuous function \citep{shu2023learning}, which exhibits robustness in both visual and empirical evaluations of losses.
Intuitively, additive models with pre-specifying \citep{lv2018oracle,chen2020sparse,wang2021huber,wang2023tilted,yuan2023sparse} could be considered as exceptional cases of MAM with a proper weighting function.

\begin{figure*}[hb]
\centering
\subfigure[Lasso]{
\includegraphics[width=0.30\textwidth]{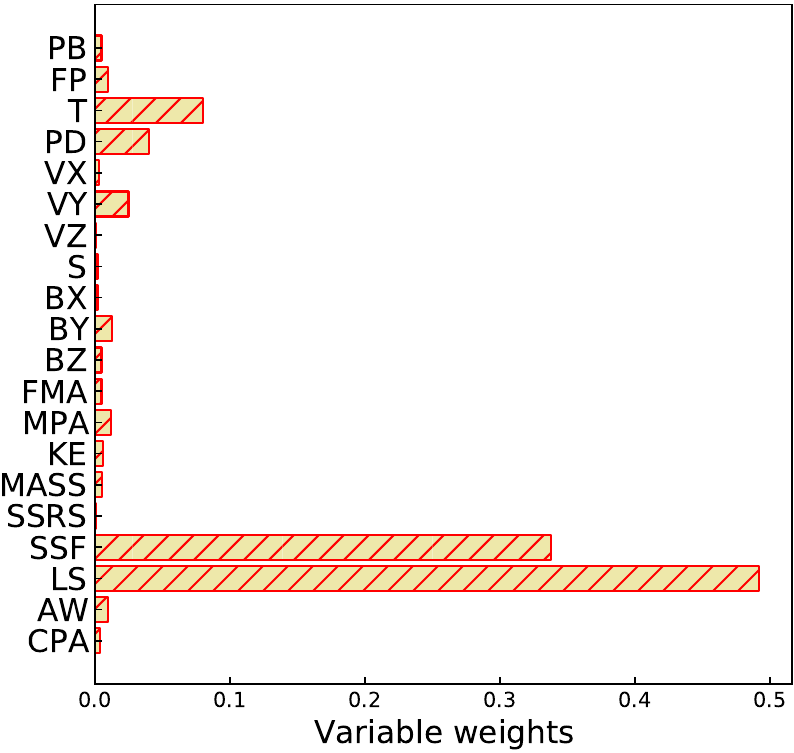}
}
\subfigure[SpAM]{
\includegraphics[width=0.30\textwidth]{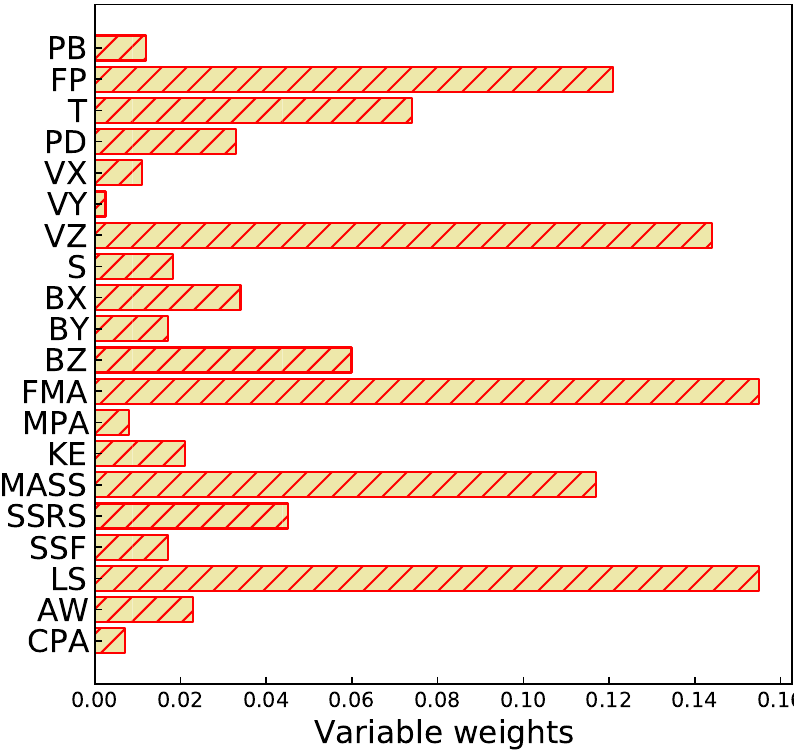}
}
\subfigure[RMR]{
\includegraphics[width=0.30\textwidth]{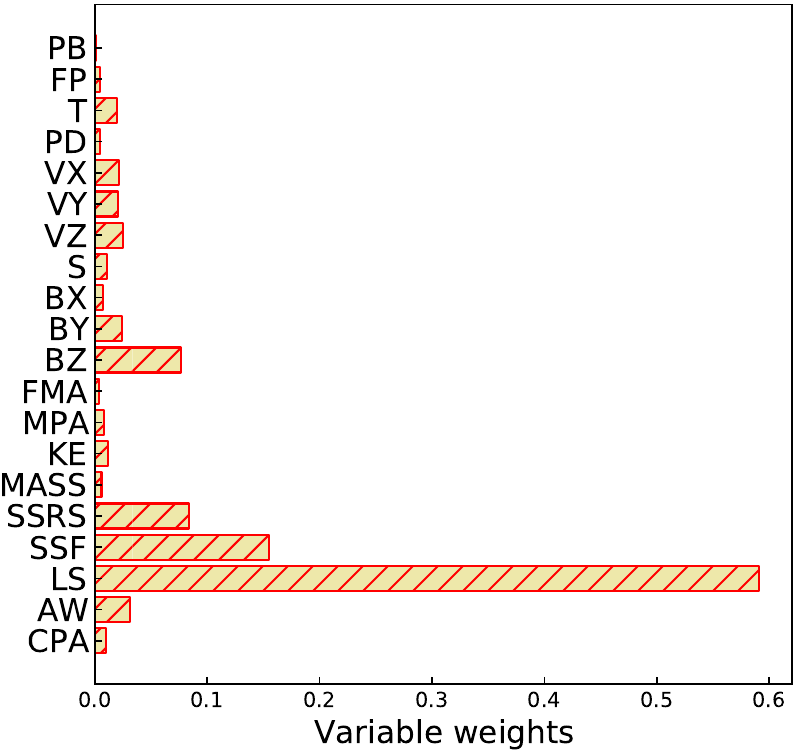}
}
\subfigure[TSpAM]{
\includegraphics[width=0.30\textwidth]{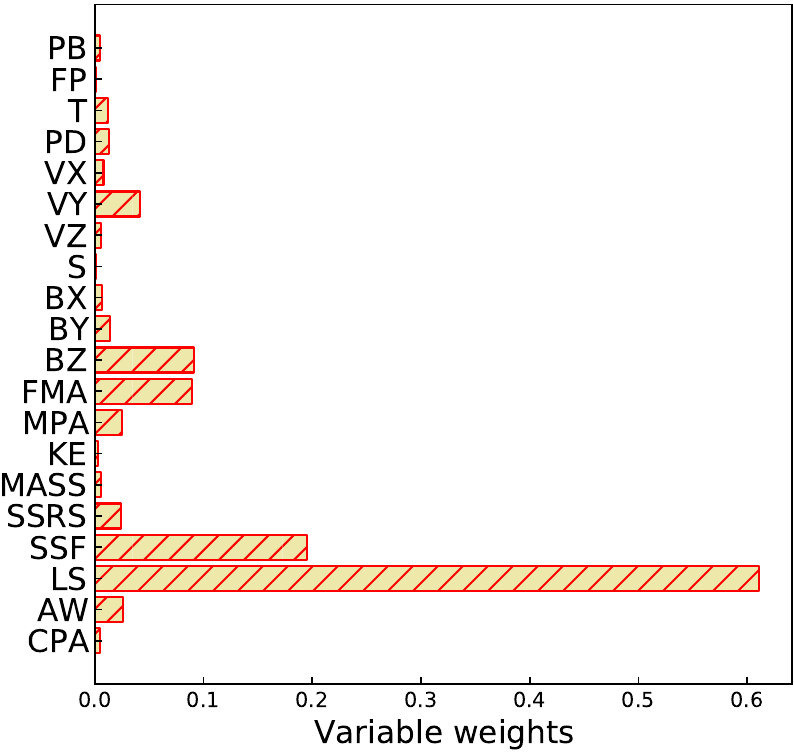}
}
\subfigure[MAM (ours)]{
\includegraphics[width=0.30\textwidth]{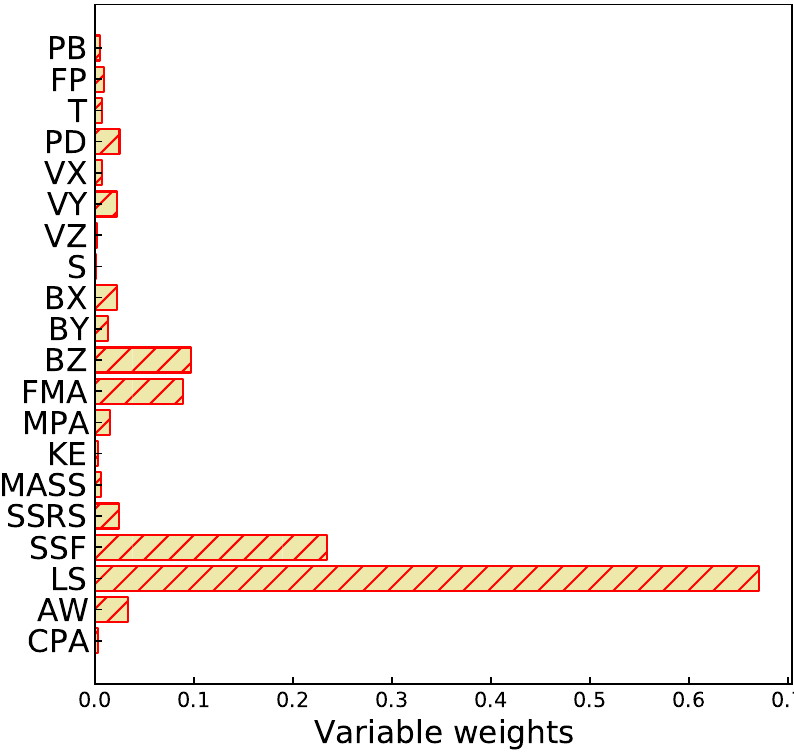}
}
\caption{The variable weights learned by several methods for the CME dataset.}
\label{f2_CME_features}
\end{figure*}

\subsection{Variable Selection on CME data}

CMEs are the most violent eruptions in the Solar System \citep{chen2011cme,veronig2021cme}. 
The relevant data are collected by combining CME and in-situ solar wind features, consisting of a single output variable (the CME arrival time) and 20 input variables. The input includes the plasma beta (PB), flow pressure (FP), temperature (T), proten density (PD), VZ, VY, VX, speed (S), BZ, BY, BX, field magnitude average (FMA), measurement position angle (MPA), kinetic energy (KE), MASS, SND speed 20RS (SSRS), SND speed final (SSF), linear speed (LS), angle width (AW), center projection angle (CPA).

Moreover, Figure \ref{f2_CME_features} shows the weights of all variables w.r.t. the arrival time prediction learned by Lasso, SpAM, RMR, TSpAM, and MAM. We observe that BZ, FMA, SSF, and LS are the significant variables identified by MAM, which is consistent with existing studies \citep{wang2023tilted,liu2018new}.

\subsection{Interpretability and Verification on Individual Features}

To rigorously evaluate the interpretability and feature disentanglement capabilities of the MAM framework, we conducted a comprehensive simulation study utilizing synthetic datasets \cite{chen2020sparse}. The mathematical formulation governing the generation of these outputs is defined as follows:
\begin{equation}
\begin{aligned}
y=f_1(x^{(1)})+ f_2(x^{(2)})+ f_3(x^{(3)})+ f_4(x^{(4)}) +\epsilon,
\end{aligned}
\end{equation}
where $f_1(x) = \ln(x), f_2(x)=e^{-x}, f_3(x)=sin(x)$ and $f_4(x)=sin(2x)$.

\begin{figure}[!t]
\centering
\subfigure[Component Functions of Ground Truth]{
\includegraphics[width=0.47\textwidth]{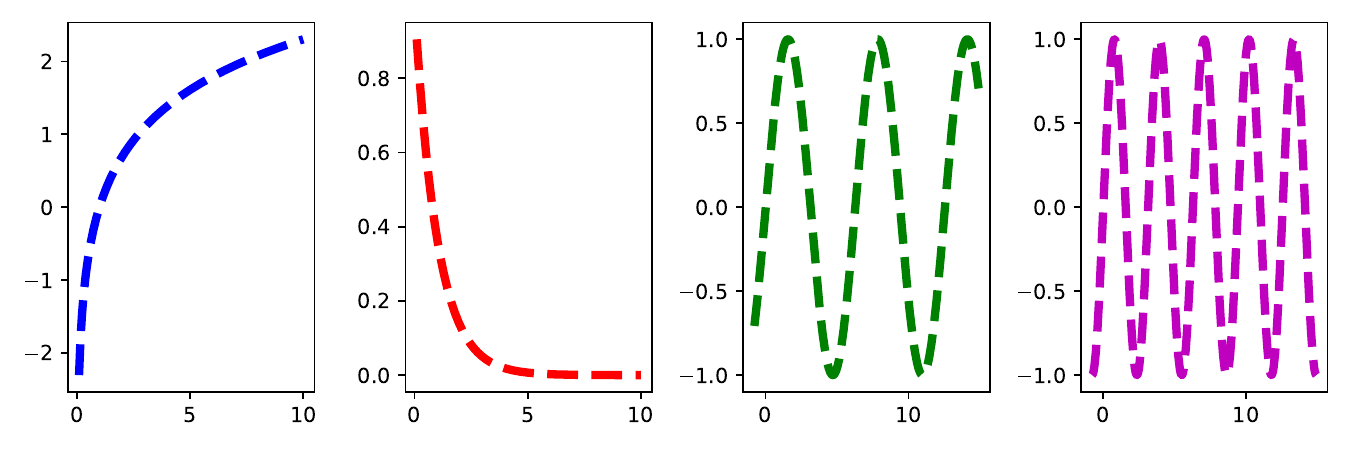}
}\\
\subfigure[With Gaussian Noise]{
\includegraphics[width=0.47\textwidth]{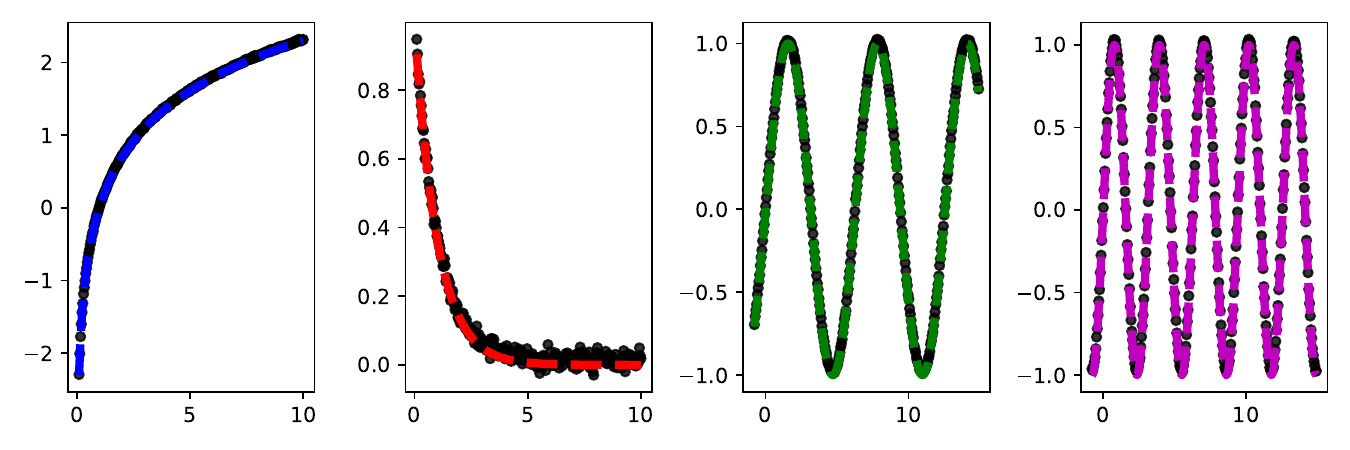}
}
\subfigure[With Gaussian Noise and $10\%$ Outliers]{
\includegraphics[width=0.47\textwidth]{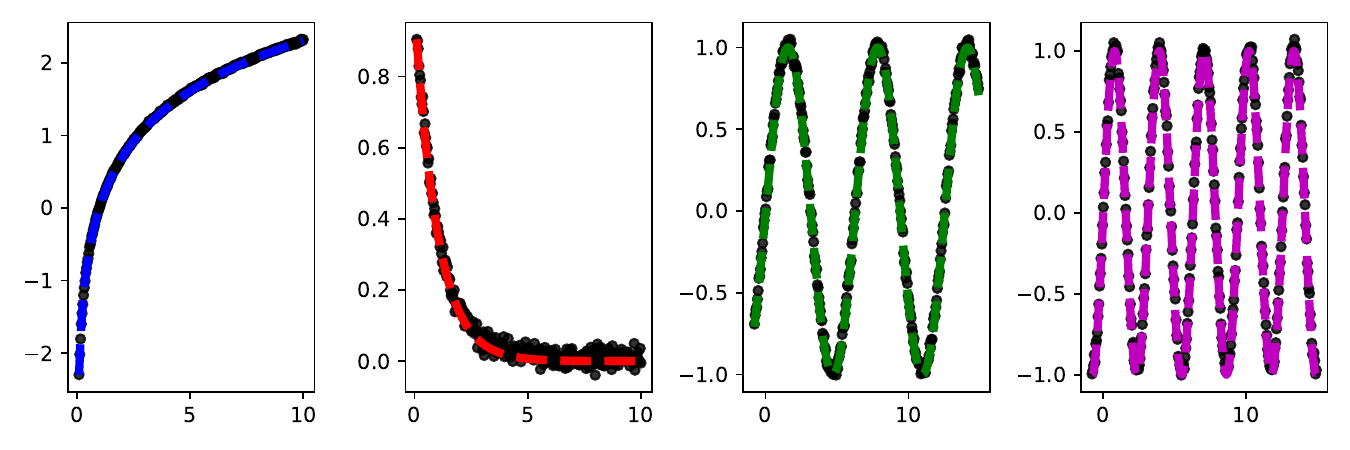}
}
\subfigure[With Cauchy Noise]{
\includegraphics[width=0.47\textwidth]{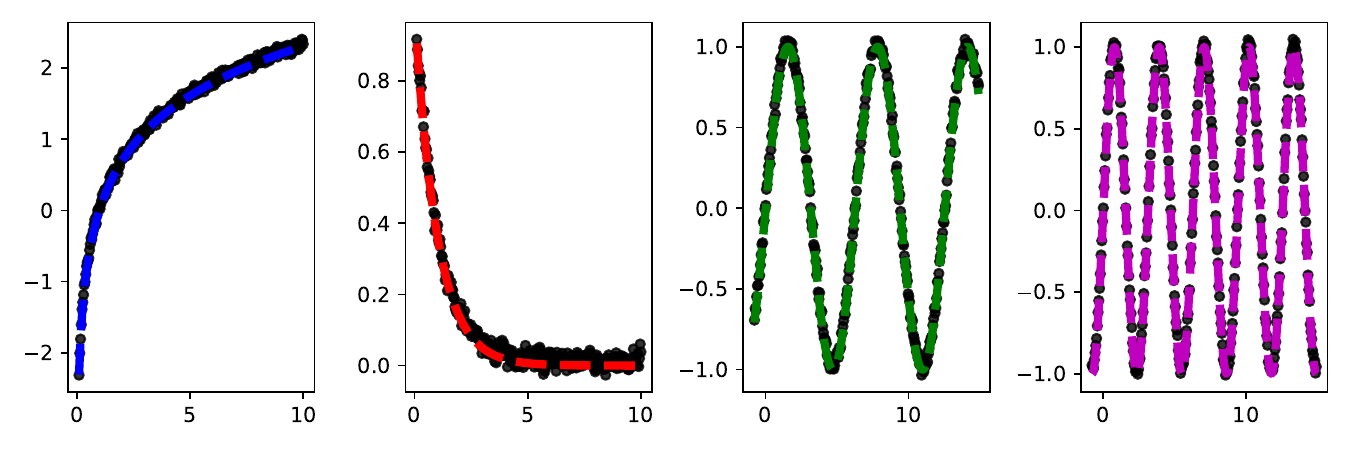}
}
\subfigure[With Cauchy Noise and $10\%$ Outliers]{
\includegraphics[width=0.47\textwidth]{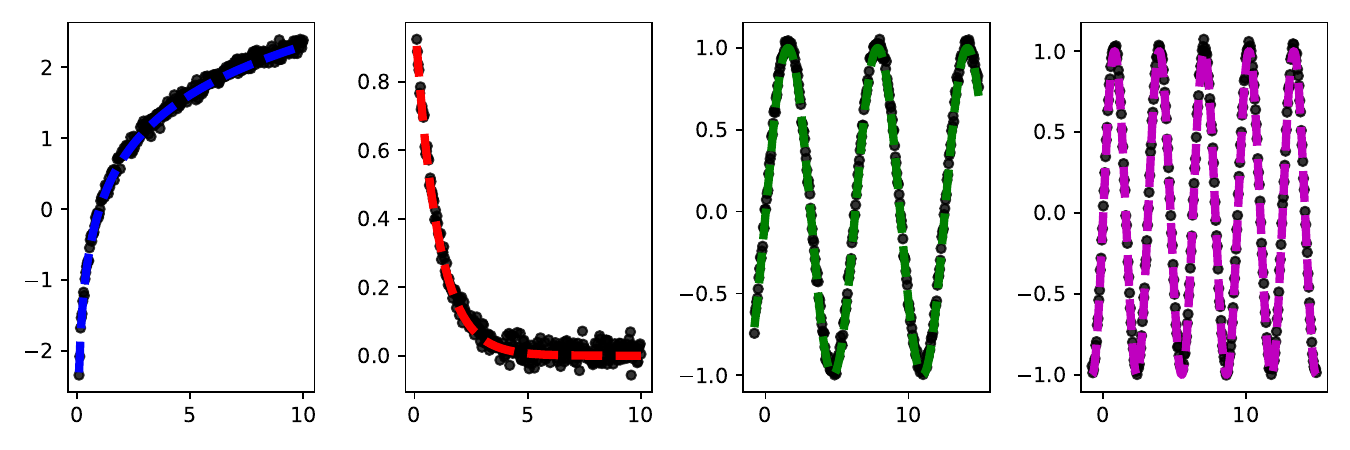}
}
\subfigure[With Student-t Noise]{
\includegraphics[width=0.47\textwidth]{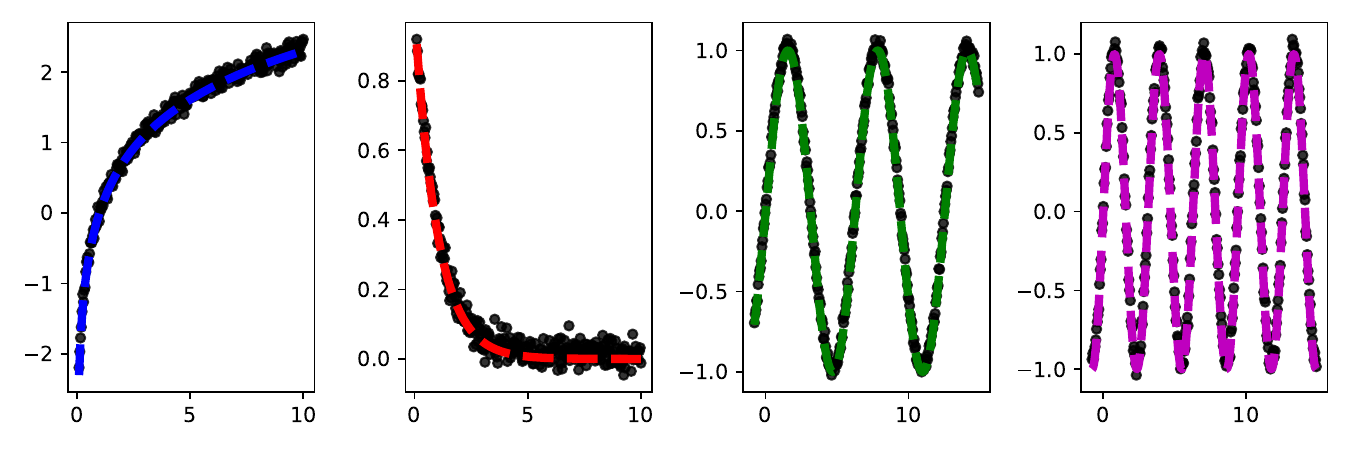}
}
\subfigure[With Student-t Noise and $10\%$ Outliers]{
\includegraphics[width=0.47\textwidth]{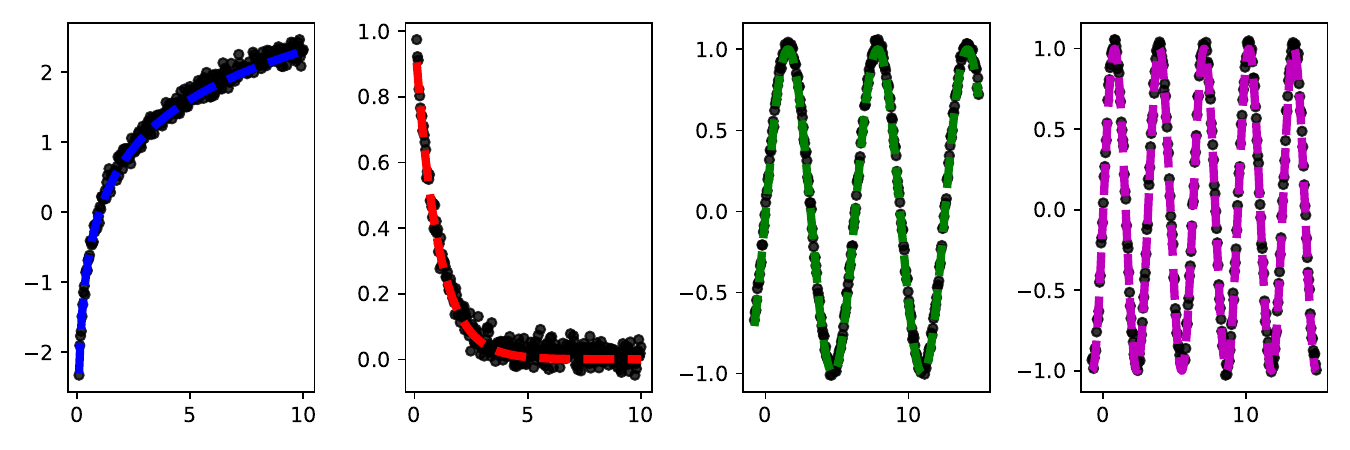}
}
\caption{Visualization of the estimated component functions by MAM. Three types of noise and the outlier are considered for validating the robustness.}
\label{fig:explainability}
\end{figure}

To assess the model's robustness against outliers and data perturbations, we consider four types of corruptions in the training dataset: (a) Normal Gaussian noise $\mathcal{N}(0,1)$; (b) Cauchy noise with freedom 2; (c) Student-t noise with freedom 2, and (d) outliers following $\mathcal{N}(100,100)$. Specifically, none or $10\%$ of the training samples were contaminated with the outliers. The explicit visualizations of the ground-truth and estimated component functions, $f_1(\cdot)$, $f_2(\cdot)$, $f_3(\cdot)$, and $f_4(\cdot)$, are provided in Figure \ref{fig:explainability}.

Figure \ref{fig:explainability} presents a qualitative comparison between the actual synthetic functions for the learning target (which integrate four distinct features) and the interpretable component functions estimated by the MAM model. The visualization demonstrates that MAM can accurately approximate the underlying ground-truth functions for individual features. This high-fidelity reconstruction is particularly significant given substantial noise in the data, suggesting that the model effectively captures the intrinsic structure of the tasks while filtering out stochastic perturbations.

\begin{table*}[htbp]
\centering
\caption{Estimation results of the component functions (average MAE ($\downarrow$) $\pm$ standard deviation). The left and right panels report the average performance with or without outliers, respectively.}
\resizebox{\textwidth}{!}{
\begin{tabular}{l|ccc|ccc}
\hline
\multirow{3}{*}{\makecell[c]{Component Functions}} 
& \multicolumn{3}{c|}{No Outliers} 
& \multicolumn{3}{c}{$10\%$ Outliers} \\
\cline{2-7}
& \makecell[c]{Gaussian\\ Noise} & \makecell[c]{Cauchy\\ Noise} & \makecell[c]{Student-t\\ Noise} & \makecell[c]{Gaussian\\ Noise} & \makecell[c]{Cauchy\\ Noise} & \makecell[c]{Student-t\\ Noise} \\
\hline
$f_1(x)=\ln(x)$             & 0.016 $\pm$ 0.004 & 0.027 $\pm$ 0.008 & 0.036 $\pm$ 0.008 & 0.018 $\pm$ 0.005 & 0.030 $\pm$ 0.010 & 0.035 $\pm$ 0.007 \\
$f_2(x)=e^{-x}$  & 0.033 $\pm$ 0.009 & 0.038 $\pm$ 0.009 & 0.054 $\pm$ 0.012 & 0.037 $\pm$ 0.010 & 0.042 $\pm$ 0.011 & 0.058 $\pm$ 0.011 \\ 
$f_3(x)=sin(x)$ & 0.019 $\pm$ 0.006 & 0.022 $\pm$ 0.006 & 0.027 $\pm$ 0.010 & 0.021 $\pm$ 0.008 & 0.028 $\pm$ 0.008 & 0.026 $\pm$ 0.011 \\
$f_4(x)=sin(2x)$ & 0.025 $\pm$ 0.007 & 0.024 $\pm$ 0.009 & 0.035 $\pm$ 0.011 & 0.028 $\pm$ 0.009 & 0.027 $\pm$ 0.012 & 0.034 $\pm$ 0.007 \\
\hline
\end{tabular}}
\label{feature_mse}
\end{table*}

Furthermore, the results in Table \ref{feature_mse} highlight MAM's ability to disentangle the feature-wise representations ($f_1,f_2,f_3,f_4$), even under various types of noise and outliers. In real-world applications where ground-truth functions are invariably latent, such precise interpretability is invaluable for validating model reliability and gaining insights into the decision-making process.

\section{Conclusion} \label{section5}
This paper proposes a novel meta-additive model (MAM) for variable selection and robust estimation. Instead of specifying a robust loss function with manually selected hyperparameters in the previous additive models, this work considers the weighting function as an additional learnable object updated by the bilevel optimization framework. For MAM, theoretical analysis assures its optimization convergence, algorithmic generalization and variable selection consistency, and experimental evaluations verify its superior performance under scenarios with complex data bias.


\section{Data availability}
Data will be made available on request.

\section{Declaration of competing interest}
The authors declare that they have no known competing financial interests or personal relationships that could have appeared to influence the work reported in this paper. 


\section{Declaration of generative AI use}
During the preparation of this work, the authors used Gemini and GLM in order to improve editing (e.g., grammar, spelling, word choice). After using this tool/service, the authors reviewed and edited the content as needed and take full responsibility for the content of the published article.

\newpage
\bibliographystyle{elsarticle-num} 
\bibliography{MAM.bib}

@inproceedings{lu2022additive,
title={Additive Gaussian processes revisited},
author={Lu, Xiaoyu and Boukouvalas, Alexis and Hensman, James},
booktitle={International Conference on Machine Learning},
pages={14358--14383},
year={2022}
}

@article{agarwal2021neural,
title={Neural additive models: Interpretable machine learning with neural nets},
author={Agarwal, Rishabh and Melnick, Levi and Frosst, Nicholas and Zhang, Xuezhou and Lengerich, Ben and Caruana, Rich and Hinton, Geoffrey E},
journal={Advances in neural information processing systems},
volume={34},
pages={4699--4711},
year={2021}
}

@article{radenovic2022neural,
title={Neural basis models for interpretability},
author={Radenovic, Filip and Dubey, Abhimanyu and Mahajan, Dhruv},
journal={Advances in Neural Information Processing Systems},
volume={35},
pages={8414--8426},
year={2022}
}

@article{fang2024generalizing,
  title={Generalizing importance weighting to a universal solver for distribution shift problems},
  author={Fang, Tongtong and Lu, Nan and Niu, Gang and Sugiyama, Masashi},
  journal={Advances in Neural Information Processing Systems },
  volume={36},
  year={2024}
}

@article{liu2025kan,
title={KAN: Kolmogorov–Arnold Networks},
author={Liu, Ziming and Wang, Yixuan and Vaidya, Sachin and Ruehle, Fabian and Halverson, James and Solja{\v{c}}i{\'c}, Marin and Hou, Thomas Y and Tegmark, Max},
booktitle={International Conference on Learning Representations},
year={2022}
}

@inproceedings{duong2024cat,
title={CAT: Interpretable Concept-based Taylor Additive Models},
author={Duong, Viet and Wu, Qiong and Zhou, Zhengyi and Zhao, Hongjue and Luo, Chenxiang and Zavesky, Eric and Yao, Huaxiu and Shao, Huajie},
booktitle={Proceedings of the 30th ACM SIGKDD Conference on Knowledge Discovery and Data Mining},
pages={723--734},
year={2024}
}

@inproceedings{franceschi2017forward,
title={Forward and reverse gradient-based hyperparameter optimization},
author={Franceschi, Luca and Donini, Michele and Frasconi, Paolo and Pontil, Massimiliano},
booktitle={International Conference on Machine Learning},
pages={1165--1173},
year={2017}
}

@article{christmann2016learning,
  title={Learning rates for the risk of kernel-based quantile regression estimators in additive models},
  author={Christmann, Andreas and Zhou, Ding-Xuan},
  journal={Analysis and Applications},
  volume={14},
  number={03},
  pages={449--477},
  year={2016},
  publisher={World Scientific}
}

@article{chen2020sparse,
  title={Sparse modal additive model},
  author={Chen, Hong and Wang, Yingjie and Zheng, Feng and Deng, Cheng and Huang, Heng},
  journal={IEEE Transactions on Neural Networks and Learning Systems},
  volume={32},
  number={6},
  pages={2373--2387},
  year={2020},
  publisher={IEEE}
}

@inproceedings{wang2017regularized,
title = {Regularized Modal Regression with Applications in Cognitive Impairment Prediction},
author = {Wang, Xiaoqian and Chen, Hong and Cai, Weidong and Shen, Dinggang and Huang, Heng},
booktitle = {Advances in Neural Information Processing Systems},
pages={1448--1458},
year = {2017}
}

@article{shi2013learning,
title={Learning theory estimates for coefficient-based regularized regression},
author={Shi, Lei},
journal={Appl. Comput. Harmon. Anal.},
volume={34},
number={2},
pages={252--265},
year={2013},
publisher={Elsevier}
}

@inproceedings{lei2023generalization,
title={Generalization analysis for contrastive representation learning},
author={Lei, Yunwen and Yang, Tianbao and Ying, Yiming and Zhou, Ding-Xuan},
booktitle={International Conference on Machine Learning},
pages={19200--19227},
year={2023}
}

@article{39,
author  = { Li, Zhu and Ton, Jean-Francois  and  Oglic, Dino and Sejdinovic, Dino },
title   = {Towards a Unified Analysis of Random Fourier Features},
journal = {J. Mach. Learn. Res.},
year    = {2021},
volume  = {22},
number  = {108},
pages   = {1-51},
}

@inproceedings{kandasamy2016additive,
  title={Additive approximations in high dimensional nonparametric regression via the SALSA},
  author={Kandasamy, Kirthevasan and Yu, Yaoliang},
  booktitle={International conference on machine learning},
  pages={69--78},
  year={2016},
  organization={PMLR}
}

@article{lv2018oracle,
title={Oracle inequalities for sparse additive quantile regression in reproducing kernel Hilbert space},
author={Lv, Shaogao and Lin, Huazhen and Lian, Heng and Huang, Jian},
journal={The Annals of Statistics},
volume={46},
number={2},
pages={781--813},
year={2018},
publisher={JSTOR}
}

@article{raskutti2012minimax,
  title={Minimax-optimal rates for sparse additive models over kernel classes via convex programming},
  author={Raskutti, Garvesh and Wainwright, Martin J and Yu, Bin},
  journal={The Journal of Machine Learning Research},
  volume={13},
  number={1},
  pages={389--427},
  year={2012},
  publisher={JMLR. org}
}

@article{stone1985additive,
  title={Additive regression and other nonparametric models},
  author={Stone, Charles J},
  journal={The annals of Statistics},
  pages={689--705},
  year={1985},
  publisher={JSTOR}
}

@inproceedings{li2020tilted,
title={Tilted Empirical Risk Minimization},
author={Tian Li and Ahmad Beirami and Maziar Sanjabi and Virginia Smith},
booktitle={International Conference on Learning Representations},
year={2021},
}

@article{tib1994lasso,
title={Regression Shrinkage and Selection Via the Lasso},
author={Tibshirani, Robert},
journal={Journal of the Royal Statistical Society, Series B},
volume={73},
number={3},
pages={267-288},
year={1994}
}

@article{hu2013learning,
title={Learning Theory Approach to Minimum Error Entropy Criterion},
author={Hu, Ting and Fan, Jun and Wu, Qiang and Zhou, Ding-Xuan},
journal={J. Mach. Learn. Res.},
volume={14},
pages={377--397},
year={2013},
publisher={Citeseer}
}

@article{fan2016consistency,
title={Consistency analysis of an empirical minimum error entropy algorithm},
author={Fan, Jun and Hu, Ting and Wu, Qiang and Zhou, Ding-Xuan},
journal={Appl. Comput. Harmon. Anal.},
volume={41},
number={1},
pages={164--189},
year={2016},
publisher={Elsevier}
}

@incollection{thrun1998lifelong,
title={Lifelong learning algorithms},
author={Thrun, Sebastian},
booktitle={Learning to learn},
pages={181--209},
year={1998},
publisher={Springer}
}

@article{vilalta2002perspective,
title={A perspective view and survey of meta-learning},
author={Vilalta, Ricardo and Drissi, Youssef},
journal={Artif. Intell. Rev.},
volume={18},
pages={77--95},
year={2002},
publisher={Springer}
}

@inproceedings{zhao2012sam,
title={Sparse additive machine},
author={Zhao, Tuo and Liu, Han},
booktitle={Artificial Intelligence and Statistics},
pages={1435--1443},
year={2012},
organization={PMLR}
}

@article{16,
author = { Tan, Zhiqiang and  Zhang, Cun-Hui},
title = {Doubly penalized estimation in additive regression with high-dimensional data},
volume = {47},
journal = {Ann. Stat.},
number = {5},
pages = {2567 -- 2600},
year = {2019},
}

@inproceedings{zhou2022core,
title={Probabilistic bilevel coreset selection},
author={Zhou, Xiao and Pi, Renjie and Zhang, Weizhong and Lin, Yong and Chen, Zonghao and Zhang, Tong},
booktitle={International Conference on Machine Learning},
pages={27287--27302},
year={2022}
}

@article{zhu2003norm,
title={1-norm support vector machines},
author={Zhu, Ji and Rosset, Saharon and Tibshirani, Robert and Hastie, Trevor},
journal={Advances in Neural Information Processing Systems (NIPS)},
volume={16},
year={2003}
}

@inproceedings{wang2021huber,
  title={Huber additive models for non-stationary time series analysis},
  author={Wang, Yingjie and Zhong, Xianrui and He, Fengxiang and Chen, Hong and Tao, Dacheng},
  booktitle={International conference on learning representations},
  year={2021}
}

@inproceedings{ji2021bilevel,
title={Bilevel optimization: Convergence analysis and enhanced design},
author={Ji, Kaiyi and Yang, Junjie and Liang, Yingbin},
booktitle={International Conference on Machine Learning},
pages={4882--4892},
year={2021}
}

@article{liang2023lower,
title={Lower bounds and accelerated algorithms for bilevel optimization},
author={Liang, Yingbin and others},
journal={J. Mach. Learn. Res.},
volume={24},
number={22},
pages={1--56},
year={2023}
}

@article{chen2021sparse,
  title={Sparse additive machine with ramp loss},
  author={Chen, Hong and Guo, Changying and Xiong, Huijuan and Wang, Yingjie},
  journal={Analysis and Applications},
  volume={19},
  number={03},
  pages={509--528},
  year={2021},
  publisher={World Scientific}
}

@article{yuan2023sparse,
  title={Sparse additive machine with the correntropy-induced loss},
  author={Yuan, Peipei and You, Xinge and Chen, Hong and Wang, Yingjie and Peng, Qinmu and Zou, Bin},
  journal={IEEE Transactions on Neural Networks and Learning Systems},
  volume={36},
  number={2},
  pages={1989--2003},
  year={2023},
  publisher={IEEE}
}

@article{shu2019meta,
title={Meta-weight-net: Learning an explicit mapping for sample weighting},
author={Shu, Jun and Xie, Qi and Yi, Lixuan and Zhao, Qian and Zhou, Sanping and Xu, Zongben and Meng, Deyu},
journal={Advances in Neural Information Processing Systems (NeurIPS)},
volume={32},
year={2019}
}

@article{shu2023cmw,
  title={Cmw-net: Learning a class-aware sample weighting mapping for robust deep learning},
  author={Shu, Jun and Yuan, Xiang and Meng, Deyu and Xu, Zongben},
  journal={IEEE Transactions on Pattern Analysis and Machine Intelligence},
  volume={45},
  number={10},
  pages={11521--11539},
  year={2023},
  publisher={IEEE}
}

@article{shu2023learning,
  title={Learning an explicit hyper-parameter prediction function conditioned on tasks},
  author={Shu, Jun and Meng, Deyu and Xu, Zongben},
  journal={Journal of machine learning research},
  volume={24},
  number={186},
  pages={1--74},
  year={2023}
}

@article{zhao2021probabilistic,
  title={A probabilistic formulation for meta-weight-net},
  author={Zhao, Qian and Shu, Jun and Yuan, Xiang and Liu, Ziming and Meng, Deyu},
  journal={IEEE Transactions on Neural Networks and Learning Systems},
  volume={34},
  number={3},
  pages={1194--1208},
  year={2021},
  publisher={IEEE}
}

@inproceedings{wang2023tilted,
  title={Tilted sparse additive models},
  author={Wang, Yingjie and Chen, Hong and Liu, Weifeng and He, Fengxiang and Gong, Tieliang and Fu, Youcheng and Tao, Dacheng},
  booktitle={International conference on machine learning},
  pages={35579--35604},
  year={2023},
  organization={PMLR}
}

@article{lahiri2016forward,
title={Forward stagewise additive model for collaborative multiview boosting},
author={Lahiri, Avisek and Paria, Biswajit and Biswas, Prabir Kumar},
journal={IEEE Transactions on Neural Networks and Learning Systems},
volume={29},
number={2},
pages={470--485},
year={2016},
publisher={IEEE}
}

@article{chen2011cme,
title={Coronal mass ejections: models and their observational basis},
author={Chen, PF},
journal={Living Reviews in Solar Physics},
volume={8},
number={1},
pages={1--92},
year={2011},
publisher={Springer}
}

@article{veronig2021cme,
title={Indications of stellar coronal mass ejections through coronal dimmings},
author={Veronig, Astrid M and Odert, Petra and Leitzinger, Martin and Dissauer, Karin and Fleck, Nikolaus C and Hudson, Hugh S},
journal={Nature Astronomy},
volume={5},
number={7},
pages={697--706},
year={2021},
publisher={Nature Publishing Group UK London}
}

@article{liu2007spam,
  title={SpAM: Sparse additive models},
  author={Liu, Han and Wasserman, Larry and Lafferty, John and Ravikumar, Pradeep},
  journal={Advances in Neural Information Processing Systems},
  volume={20},
  year={2007}
}

@article{bao2024robust,
title={Robust embedding regression for semi-supervised learning},
author={Bao, Jiaqi and Kudo, Mineichi and Kimura, Keigo and Sun, Lu},
journal={Pattern Recognit.},
volume={145},
pages={109894},
year={2024},
publisher={Elsevier}
}

@inproceedings{DBLP:conf/nips/BaoWLZZ21,
author    = {Fan Bao and
Guoqiang Wu and
Chongxuan Li and
Jun Zhu and
Bo Zhang},
title     = {Stability and generalization of bilevel programming in hyperparameter optimization},
booktitle = {Advances in Neural Information Processing Systems},
pages     = {4529--4541},
year      = {2021}
}

@article{hoffer2017train,
title={Train longer, generalize better: closing the generalization gap in large batch training of neural networks},
author={Hoffer, Elad and Hubara, Itay and Soudry, Daniel},
journal={Advances in Neural Information Processing Systems},
volume={30},
year={2017}
}

@inproceedings{zhang2024sbo,
title     = {Fine-grained Analysis of Stability and Generalization for Stochastic Bilevel Optimization},
author    = {Zhang, Xuelin and Chen, Hong and Gu, Bin and Gong, Tieliang and Zheng, Feng},
booktitle = {International Joint Conference on Artificial Intelligence},
pages     = {5508--5516},
year      = {2024}
}

@inproceedings{lee2020maskgan,
title={Maskgan: Towards diverse and interactive facial image manipulation},
author={Lee, Cheng-Han and Liu, Ziwei and Wu, Lingyun and Luo, Ping},
booktitle={Proceedings of the IEEE/CVF conference on computer vision and pattern recognition},
pages={5549--5558},
year={2020}
}

@article{lecun1998mnist,
title={The MNIST database of handwritten digits},
author={LeCun, Yann},
journal={http://yann. lecun. com/exdb/mnist/},
year={1998}
}

@article{liu2018new,
title={A new tool for CME arrival time prediction using machine learning algorithms: CAT-PUMA},
author={Liu, Jiajia and Ye, Yudong and Shen, Chenglong and Wang, Yuming and Erd{\'e}lyi, Robert},
journal={The Astrophysical Journal},
volume={855},
number={2},
pages={109},
year={2018},
publisher={IOP Publishing}
}
\end{document}